\documentclass{article}

\PassOptionsToPackage{numbers,sort&compress}{natbib}
\usepackage{PRIMEarxiv}

\usepackage[utf8]{inputenc} 
\usepackage[T1]{fontenc}    
\usepackage{hyperref}       
\usepackage{url}            
\usepackage{booktabs}       
\usepackage{amsfonts}       
\usepackage{nicefrac}       
\usepackage{microtype}      
\usepackage{lipsum}
\usepackage[table]{xcolor} 
\usepackage{fancyhdr}       
\usepackage{graphicx}       
\graphicspath{{media/}}     
\usepackage{amsmath} 
\usepackage{multirow}
\usepackage{caption}

\title{Causal Disentanglement Learning for Accurate Anomaly Detection in Multivariate Time Series}

\author{
  Wonah Kim, Jeonghyeon Park, Dongsan Jun, Jungkyu Han,  Sejin Chun \\
  Dong-A University \\
  \texttt{2243742@donga.ac.kr, jeonghyeon.park@datasciencelabs.org, \{dsjun,jkhan,sjchun\}@dau.ac.kr} \\
}

\begin{document}
\maketitle

\begin{abstract}
 Disentangling complex causal relationships is important for accurate detection of anomalies. In multivariate time series analysis, dynamic interactions among data variables over time complicate the interpretation of causal relationships. Traditional approaches assume statistical independence between variables in unsupervised settings, whereas recent methods capture feature correlations through graph representation learning. However, their representations fail to explicitly infer the causal relationships over different time periods. To solve the problem, we propose Causally Disentangled Representation Learning for Anomaly Detection (CDRL4AD) to detect anomalies and identify their causal relationships in multivariate time series. First, we design the causal process as model input, the temporal heterogeneous graph, and causal relationships. Second, our representation identifies causal relationships over different time periods and disentangles latent variables to infer the corresponding causal factors. Third, our experiments on real-world datasets demonstrate that CDRL4AD outperforms state-of-the-art methods in terms of accuracy and root cause analysis. Fourth, our model analysis validates hyperparameter sensitivity and the time complexity of CDRL4AD. Last, we conduct a case study to show how our approach assists human experts in diagnosing the root causes of anomalies.
\end{abstract}


\section{Introduction}
Anomaly detection is critical for timely decision-making in safety-sensitive domains, such as cybersecurity \cite{HAI}, server performance monitoring \cite{jia2023robust}, and failure prediction and prevention \cite{OmniAnomaly}. In cybersecurity, effective anomaly detection involve identifying irregular network activities, unauthorized access, or atypical user behaviors \cite{deviation}. However, complex interactions among data variables complicate anomaly detection. This emphasizes the necessity to explicitly understand both causal and correlational relationships \cite{causality_ad}. Thus, advanced modeling techniques that explicitly disentangle underlying causal relationships are essential.

Building upon the foundational principles, the importance of Multivariate Time Series (MTS) analysis becomes especially evident \cite{MARINA}. MTS involve multiple interrelated variables observed over time, posing additional complexities to anomaly detection tasks. Conventional methods\cite{LSTM-NDT, LSTM-VAE} assume that data is \textit{independent and identically distributed} (aka \textit{i.i.d.}). These methods suffer from performance degradation in multivariate contexts where associations between variables are crucial. They fail to effectively accommodate dependencies and interactions. This leads to suboptimal anomaly detection across various time series. Consequently, there is a pressing need to elucidate the intricate relationships in MTS data.

Recently, Graph Neural Networks (GNNs) have gained popularity due to the fact that they capture interactions between variables in anomaly detection \cite{GNN_AD, BWGNN, gnnAD_tran}. Typically, GNN-based methods identify anomalies as unusual \textit{nodes} within graph structures. They leverage feature representations that encapsulate structural and attribute information. By aggregating features from a node's local neighborhood, GNNs learn complex interactions that signal potential anomalies and effectively capture contextual information.

Existing GNN studies\cite{MTAD-GAT, DuoGAT, FuSAGNet} primarily focus on temporal dependencies and associations between variables. However, they do not explicitly distinguish cause-and-effect relationships. Approaches such as MTAD-GAT \cite{MTAD-GAT} and DUO-GAT \cite{DuoGAT} model feature correlations alongside temporal dependencies across different time periods. FuSAGNet \cite{FuSAGNet} employs a Bidirectional Gated Recurrent Unit to capture temporal dynamics. Still, these methods do not explicitly incorporate domain-specific causal structures. This limits their ability to detect anomalies accurately in MTS contexts.

Causal Representation Learning (CRL) addresses this limitation by explicitly encoding causal relationships. Existing CRL methods \cite{CausalVAE, CR-VAE, CausalGNN, GANF} predominantly focus on determining causal relationships among past events. They use mechanisms like causal graphs \cite{CausalVAE} or multi-head decoders \cite{CR-VAE}.  However, these methods often neglect the varying significance of events across different time periods. They assume instantaneous rather than gradual influences, thus inadequately capturing time-lagged causal relationships (\textit{aka} Granger causality \cite{Granger}). This assumption limits their effectiveness and can lead to suboptimal decisions due to incorrect causal assumptions.

Disentangled Representation Learning (DRL) has attracted significant interest over the past decade. DRL decomposes high-dimensional data into interpretable latent variables representing independent factors of data variation. Notable approaches \cite{SS-FVAE-BB, CaD-VAE} aim to achieve statistical independence between latent variables. Other DRL studies encode biases \cite{DisC}, temporal variations \cite{S3VAE}, and multi-view contexts \cite{Multi-VAE}. Nevertheless, these methods typically do not explicitly distinguish causal relationships between distinct factors. This limitation reduces their effectiveness in accurately detecting anomalies within the underlying causal structures.

To address these challenges, we propose Causally Disentangled Representation Learning for Anomaly Detection (\textbf{CDRL4AD}). Our method explicitly disentangles complex causal relationships for anomaly detection in MTS. Our key contributions include:

\begin{itemize}
    \item We introduce a comprehensive causal representation framework within \textbf{CDRL4AD}. This framework integrates rich graph structures to address the heterogeneity, temporal dynamics, and causality inherent in MTS. To the best of our knowledge, no prior work has simultaneously addressed these three aspects  for accurate anomaly detection within intricate causal networks.

    \item Our causally disentangled representations capture time-lagged causal relationships through causal discovery. Latent variables are then encoded into interpretable aspects of time series related to causal mechanisms. Then, our correlation representation at both node and edge levels, along with temporal dependency modeling. Thus, we embed causal discovery, disentangled representation learning (DRL), and correlation and temporal dependency modeling into a unified GNN framework. 
    
    \item We demonstrate through extensive evaluation that our model outperforms state-of-the-art methods on real-world datasets regarding accuracy and root cause analysis. Additionally, we present an in-depth analysis focusing on performance overhead, and sensitivity to hyperparameters. Lastly, we provide a case study in which the root causes of anomalies are diagnosed by humans.
\end{itemize}

The rest of this paper is organized as follows. Section \ref{related} reviews the literature on anomaly detection under MTS context, as well as causal and disentanglement learning methods. We first formulate our problem and the causal process, and then present CDRL4AD in Section \ref{proposed}. In Section \ref{evaluation}, we discuss the experimental results and analyze our model in detail. Next, we showcase a case study of anomaly diagnosis examined by humans, in Section \ref{case_study}. Finally, Section \ref{conclusion} concludes our research.

\section{Related Work} \label{related}

In this section, we first review conventional approaches to anomaly detection in MTS. Next, we explore GNN-based models for detecting anomalies in MTS. Lastly, recent studies on causal representation learning are investigated.

\subsection{Anomaly Detection in MTS}
Over the past decade, deep learning models for anomaly detection have primarily been classified into prediction-based and reconstruction-based approaches. Prediction-based models detect anomalies by comparing actual values with predicted outcomes. For instance, AD-LTI \cite{AD-LTI} utilizes Prophet and a GRU network, while LSTM-NDT \cite{LSTM-NDT} employs an LSTM model to forecast anomalies based on threshold deviations.

Reconstruction-based models identify anomalies through reconstruction errors from latent variables. OmniAnomaly \cite{OmniAnomaly} computes reconstruction errors by learning data representation through stochastic RNN and planar normalization flow. Moreover, both USAD \cite{USAD} and DAGMM \cite{DAGMM} employ autoencoders to improve performance, while MSCRED \cite{MSCRED} models both feature correlation and temporal dependency using convolutional encoder-decoder. To handle temporal dependencies, MAD-GAN \cite{MAD-GAN} and TranAD \cite{TranAD} leverage generator and discriminator networks based on both GAN and LSTM-RNN, and transformer, respectively. More recently, IMDIFFUSION \cite{IMDIFFUSION} adopts a combination of diffusion and transformer to reconstruct masked data for anomaly score calculation. However, their models lack the topological relationships necessary to express which variables are important. Consequently, they fail to effectively capture the dynamic evolution of interactions among the variables over time.

\subsection{GNNs for Detecting Anomalies}
Graph Neural Networks (GNNs) \cite{GNN} enhance feature representation by incorporating the topological relationships into graph representation learning. In GNNs, each node updates its features by integrating the features of its adjacent nodes. The node features are updated by aggregating neighboring node features, utilizing aggregation functions such as sum, average, and max. The updated information is amalgamated with the original node features for subsequent updates after the aggregation. 

Two prevalent mechanisms of updating features in GNNs are Graph Convolution Networks (GCNs) \cite{GCN} and Graph Attention Networks (GATs) \cite{GAT}. GCNs update node features by averaging neighboring features followed by a linear transformation. In contrast, GATs use attention mechanisms to assign dynamic weights to neighboring nodes, prioritizing the most pertinent neighbor information.

In the context of anomaly detection, several studies have highlighted the advantages of GNN-based representation learning \cite{GDN, MTAD-GAT, FuSAGNet, GReLeN}. Their representations encode both correlations and temporal dependencies while distinguishing the significance of individual nodes. Especially, GDN \cite{GDN} uses GAT to enhance node representations and predicts future observations based on the patterns it has learned. MTAD-GAT \cite{MTAD-GAT} models both correlations and temporal dependencies simultaneously using GAT. Similarly, FuSAGNet \cite{FuSAGNet} leverages GAT to reduce prediction errors by integrating node features with sparse latent representations learned via BiGRU. Additionally, GReLeN \cite{GReLeN} utilizes an encoder-decoder structure with GCN to effectively capture temporal dependencies.

Still, existing GNN-based methods \cite{GDN, MTAD-GAT, FuSAGNet, GReLeN} have two notable limitations. First, their representations fail to capture the full extent of the correlations within the entire graph structure. Especially, their representations often overlook anomalies in edges that can be detected through changes in correlations between nodes and edges. Second, the attention mechanisms are constrained to a single time period when assigning varying significance to nodes. These problems make it harder to model cause-and-effect relationships across different time periods.

\subsection{CRL and DRL} \label{crldrl}
Existing CRL studies face significant challenges in addressing time-lagged causal relationships. For example, CausalVAE \cite{CausalVAE} encodes causal relationships using masking mechanisms in causal graphs and decodes latent variables to minimize reconstruction errors. CR-VAE \cite{CR-VAE}, on the other hand, adopts an RNN-based encoder-decoder to identify historical data segments that influence current observations. CausalGNN \cite{CausalGNN} uses GCN in an encoder-decoder architecture to capture both cause-and-effect relationships and temporal dependencies. GANF \cite{GANF} captures static causal representations through a combination of RNNs and GCNs under the assumption that causality remains consistent over time. However, these models often struggle to handle unexpected changes effectively because they do not adequately distinguish the varying significance of causal relationships over time. 
Consequently, it may lead to suboptimal decisions based on incorrect causal assumptions. 

Recently, a few studies on DRL \cite{SS-FVAE-BB, CausalVAE, CaD-VAE} have focused on decomposing high-dimensional data into interpretable latent variables, where each variable corresponds to independent factors of data variation. Specifically, SS-FVAE-BB \cite{SS-FVAE-BB} reconstructs the data encoded by mapping each factor to a unique latent variable independently. Similarly, both CausalVAE \cite{CausalVAE} and CaD-VAE \cite{CaD-VAE} present disentanglement learning, where each latent dimension is assigned to only relative factors. 

On the other hand, some studies (DISC \cite{DisC}, Multi-VAE \cite{Multi-VAE}, and S3VAE \cite{S3VAE}) incorporate a specific structure of harmonizing factors such as bias\cite{DisC}, temporal changes\cite{S3VAE}, and multi-view settings\cite{Multi-VAE}. In particular, DISC \cite{DisC} disentangles causal and bias graphs and learns embeddings using GNN modules to generate a disentangled representation. Multi-VAE \cite{Multi-VAE} disentangles images from various views into shared and unique variables for each view using reparameterization. S3VAE \cite{S3VAE} separates latent variables into a static and a dynamic representations to model temporal dynamics. None of these approaches, however, focus on aligning disentangled representations with explicit causal relationships.

\begin{figure*}[t]
    \centering
    \includegraphics[width=\textwidth]{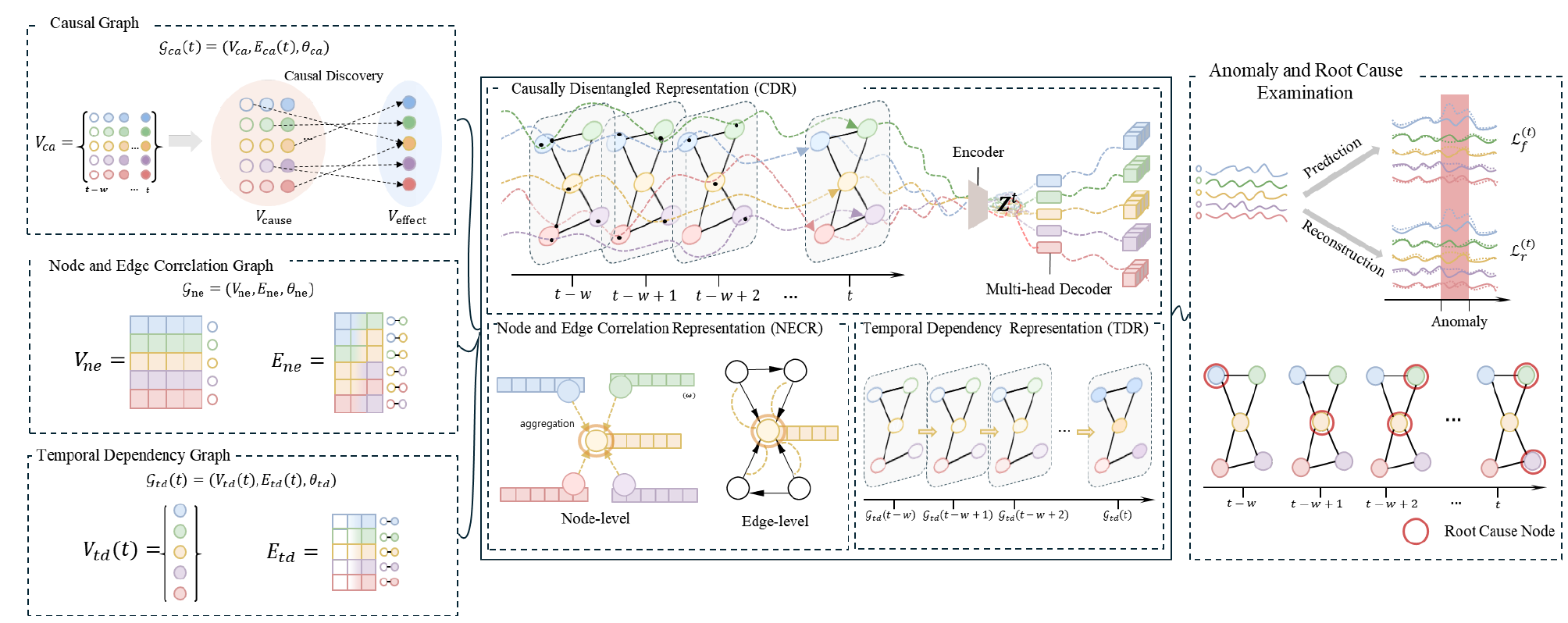}
    \caption{The overview of our proposed framework, CDRL4AD}
    \label{fig:1}
\end{figure*}

\section{Proposed Model} 
\label{proposed}

In this section, we formulate the problem and introduce the overview of our model. Then, we present our model in detail.

\subsection{Problem Formulation}\label{problem_form} 
Our multivariate time series~(MTS) is represented as a sequence of data points observed from \( N \) sensors over \( T \) timestamps, denoted by \( \textbf{X} \in \mathbb{R}^{N \times T} \). The data collected from \( N \) sensors at a specific time \( t \) is represented as \(X^{t} = \{ \textbf{x}_{1,t}, \ldots, \textbf{x}_{N,t} \} \), where \( \textbf{x}_{i,t} \in \mathbb{R}^m \) denotes the multivariate time series vector observed from the \( i \)-th sensor at time \( t \). Similarly, the data collected from \( i \)-th sensor across all timestamps is represented as \(X_{i} = \{ \textbf{x}_{i,1}, \ldots, \textbf{x}_{i,T} \} \).

To capture the temporal dependencies between the current timestamp and previous ones, we define the sample set $S^{(t)}$ at a specific time $t$ as:
\[ S^{(t)} = \{ X^{t-\omega}, ..., X^{t-1} \} \]

where $\omega$ denotes the width of the sliding window for extracting samples. This set includes all data points from the previous $\omega$ times up to immediately before time $t$~(\(t-1\)). Each element of $S^{(t)}$ is a snapshot of the system's state at a prior time. 

The goal of our anomaly detection problem is to assign a set of binary labels \( \{ 0, 1 \} \), where each label indicates whether an anomaly is detected for a given sample. Our anomaly score function \( \mathcal{A}(t) \) determines an anomalous state at time $t$ if and only if \( \mathcal{A}(t) \) exceeds a predefined causal threshold \( \Theta \). In unsupervised settings, we assume that the training data consists solely of normal states, whereas the test data contains both normal and anomalous states. This approach aims to improve detection accuracy by effectively distinguishing between normal and anomalous states based on the computed anomaly scores.

\subsection{Model Overview}
Figure \ref{fig:1} provides an overview of our proposed framework, structured into three primary phases. In the first phase, we address the inherent heterogeneity, temporal dynamics, and causal relationships in multivariate time series (MTS) data by constructing a temporal heterogeneous graph. This graph includes a node-edge correlation graph, a temporal dependency graph, and a causal graph.

The second phase focuses on Causally Disentangled Representation Learning (CDRL). This phase incorporates three distinct representations: Causally Disentangled Representation (CDR), Temporal Dependency Representation (TDR), and Node-Edge Correlation Representation (NECR). Together, these representations capture time-lagged causal relationships, temporal dependencies, and feature correlations at both node and edge levels.

In the final phase, we quantify the degree of anomaly at specific time steps by jointly optimizing prediction and reconstruction losses. Subsequently, the corresponding root-cause nodes are identified over time.

In the following section, we present the proposed model in detail with the notations in the Table~\ref{tab:1}.

\begin{table}[b]
\centering
\caption{Notations} \label{tab:1} 
    \begin{tabular}{cl}
    \hline
    \textbf{Indices}               & \textbf{Definition}                                                                               \\ \hline
    $X^t$                          & Time series at time $t$                                                                           \\
    $X_i$                          & Time series measured from variable $i$                                                            \\
    $\textbf{x}_{i,t}$             & Time series vector from variable $i$ at time $t$                                                  \\
    $S^{(t)}$                      & The set of samples at time $t$                                                                    \\
    $\mathcal{G}_{ca}(t)$          & Causal graph at time $t$                                                                          \\
    $\mathcal{G}_{td}(t)$          & Temporal dependency graph at time $t$                                                             \\
    $\mathcal{G}_{ne}$             & Node and edge correlation graph                                                                   \\
    \(\ddot{x}, \widetilde{x}\)    & Cause and effect variables of $\textbf{x}$                                                        \\
    $\alpha_{ij}^{\mathcal{R}(t)}$ & Attention weight for representation $\mathcal{R}$ between node $i$ and $j$ at time $t$ \\
    $ r_i^{(t)} $                  & Representation capturing causal relationships related to variable $i$                             \\
    $ \tilde{h}^t, \hat{h}_i^t $   & Hidden states of the encoder and $i$-th decoder head at time $t$                                  \\
    $ d^{\mathcal{R}(t)} $         & Latent dimension for representation $\mathcal{R}$ at time $t$                                     \\
    $ \mathcal{A}(t) $             & Anomaly score function at time $t$                                                                \\ \hline
    \end{tabular}
\end{table}

\subsection{Our Causal Process}
In this subsection, we provide the definition of a temporal heterogeneous graph, a time-lagged causality in variation, and a structural causal mechanism.

\subsubsection{\textbf{Temporal heterogeneous graph}}
Typically, MTS struggles to explicitly capture the complex interactions and dependencies among variables. To enrich the representation of MTS \(\textbf{X}\) and facilitate the analysis of causal relationships and complex patterns, we model it as a Temporal Heterogeneous Graph as follows.

\begin{center}
\( \mathcal{G} = ( V, E, T, \theta ) \) 
\end{center}

\noindent where  \( V \) denotes the set of $N \times |T|$ nodes \( v \in V \), \( E \) denotes the set of edges $e \in \mathbb{R}^{|V| \times |V|}$, \( T \) is the set of timestamps, and \( \theta \) is the set of weight functions \( E \rightarrow \mathbb{R} \).

Each weight function in \( \theta \) provides semantic context to the edges, capturing relationships such as causality, statistical correlation, temporal dependencies, weights, and directionality within the MTS.

Given a temporal heterogeneous graph \( \mathcal{G} \) at a specific time $t$, we define three types of subgraphs as follows:

\noindent\textbf{Causal graph.} This graph captures causal relationships from the nodes in \(\omega - 1\) consequent previous timestamps to the nodes in the current time stamp \(t\), analyzing causally influenced events for \(\omega\) samples. At a current timestamp $t$, the causal graph is defined as \( \mathcal{G}_{ca}(t, \omega) = (V_{ca}(t, \omega), E_{ca}(t, \omega), \theta_{ca}) \), where the subset \( V_{ca}(t, \omega) \subseteq V \) consists of two distinct groups: \( V_{cause} \), which includes the nodes in previous timestamps that initiating cause actions, and \( V_{effect} \), containing nodes at timestamp \(t\) that affected by the cause actions. The edge set \( E_{ca}(t, \omega) \subseteq E \) links \( u \in V_{cause} \) to \( v \in V_{effect} \), with each edge \( e \in E_{ca}(t, \omega) \) representing a direct causal link from prvious node $u$ to current node $v$ at time \( t \). The link strength of edge is calculated by the causal weight function \(\theta_{ca}\) based on the vectors \(\textbf{x}_{i,t}\) associated with individual sensor.

\noindent\textbf{Node and edge correlation graph.} This graph is essential for uncovering underlying patterns in data by highlighting significant statistical correlations among nodes and edges, inspired by \cite{NENN}. The graph is defined as \( \mathcal{G}_{ne} = (V_{ne}, E_{ne}, \theta_{ne}) \), where \( V_{ne} \) is the set of \(N\) nodes and \( E_{ne} \) is the set of edges \(e \in \mathbb{R}^{N \times N}\). Each edge represents significant statistical correlations between nodes, independent of specific timestamps. Specifically, an undirected edge \(e_{(v,v')}\) between nodes \(v\) and \(v'\) indicates the strength of the correlation, quantified through the correlation weight function \(\theta_{ne}\) that calculates cosine similarity between the embeddings of each node learned from data. Since \(\theta_{ne}\) sets the weight to 0 if the cosine similarity between the nodes \(v\) and \(v'\) are not located in the top-$K$ similar node lists of each other, the edge \(e_{(v,v')}\) demonstrates a robust correlation between \(v\) and \(v'\).

\noindent\textbf{Temporal dependency graph.} This graph captures the temporal dependencies between samples of different timestamps, explaining how relationships evolve over sequential samples. At time $t$ with sliding window size \(\omega\), the graph is defined as \( \mathcal{G}_{td}(t, \omega) = (V_{td}(t, \omega), E_{td}(\omega), \theta_{td}) \), where \( V_{td}(t, \omega)\) indicates the set of \(\omega\) nodes, each of them represents the specific samples between \(X^{t-\omega+1}\) and \(X^{t}\). Additionally, \( E_{td}(\omega)\) represents the set of edges \(e \in \mathbb{R}^{\omega \times \omega}\). \(\theta_{td}\) calculates the edge weight by examining the node embeddings learned from data. By analyzing the temporal evolution of graphs, we can monitor how changes affect the time series vectors \(X^{t-\omega+1}\) through \(X^{t}\).

\subsubsection{\textbf{A Time-Lagged Causality in Variation}} 
Time-lagged causality refers to the phenomenon where the effect of a causal action is observed after a delay, rather than immediately. Unlike the causal graph, time-lagged causality emphasizes the temporal gap between causal factor and their effects.

In the context of Granger causality\cite{Granger}, we define a \textit{time-lagged causal relationship} between variables. Granger causality is able to identify temporal dependencies, as it determines causation by assessing whether exploiting the past values of one variable improves the future value prediction of another variable.

Given two variables \(X_i\) and \(X_j\) from different sensors, \(X_j\) is said to cause \(X_i\) if the past values of \(X_j\) provide significant information about future values of \(X_i\) more than the past values of \(X_i\) alone can provide. We use $\mathrel{\cdot\cdot}$ and \( \sim \) to denote the cause and effect variables, respectively. For example, if \(X_j\) is a cause of \(X_i\), we write the variables as \(\ddot{X}_j\) and \(\widetilde{X_i}\) in the context of causal discussion. Formally, \(\ddot{X}_j\) causes \(\widetilde{X_i}\) if:
\[
    \begin{aligned}
    \text{Var}(\textbf{x}_{i,t} \mid \textbf{x}_{i,t-1}, \textbf{x}_{i,t-2}, \ldots) > 
    \text{Var}(\textbf{x}_{i,t} \mid \textbf{x}_{i,t-1}, \textbf{x}_{i,t-2}, \ldots, \textbf{x}_{j,t-1}, \textbf{x}_{j,t-2}, \ldots)
    \end{aligned}
\]
satisfied for \( \forall t \in T\), where \(\text{Var}(X|\cdot)\) denotes the conditional variance of prediction \(X\) given the specified information set \(\cdot\).

Here, we define the cause variable \( \ddot{\textbf{x}}_{j, t'} \) as a member of \(\ddot{X}_j\), identical to the time series vector \( \ddot{\textbf{x}}_{j, t'} \) of the variable $j$ at time $t'$, where $t-\omega \leq t' < t$. Likewise, \( \widetilde{X}_i \) includes the effect variable \( \widetilde{\textbf{x}}_{i, t} \) corresponding to $\textbf{x}_{i, t}$.

\subsubsection{\textbf{Latent Causal Process}} 
We derive a structural causal model (SCM) \cite{Toward_Causal, scm_tran} to capture the influence of cause variables on effect variables. The SCM is a formal framework that represents cause-and-effect relationships within the causal graph $\mathcal{G}_{ca}(t)$. Given an effect variable \(\widetilde{\textbf{x}}_{i, t}\), \( PA(\widetilde{\textbf{x}}_{i, t}) \) denotes the set of latent factors that represents the parents of \(\widetilde{\textbf{x}}_{i, t}\) and directly cause \(\widetilde{\textbf{x}}_{i, t}\).

Each relationship is governed by a local causal mechanism denoted by $ f_i (PA(\widetilde{\textbf{x}}_{i, t}),U_i) \rightarrow \widetilde{\textbf{x}}_{i, t}$, where $\widetilde{\textbf{x}}_{i, t}$, $PA(\widetilde{\textbf{x}}_{i, t})$, $U_{i}$, and \(f_i\) represent the effect variable, the set of cause variables, random noise, and a causal mechanism function, respectively. For instance, consider the variables $\textbf{x}_{i, t}, \textbf{x}_{j, t-2}$, and $\textbf{x}_{k, t-3}$ in the SCM $\textbf{x}_{j, t-2} \rightarrow \textbf{x}_{i, t} \leftarrow \textbf{x}_{k, t-3}$. The local causal mechanism for $\textbf{x}_{i, t}$ is described as $f_i(PA(\textbf{x}_{i, t}), U_i) \rightarrow \textbf{x}_{i, t} $, where $PA(\textbf{x}_{i, t})$ is $\{\textbf{x}_{j, t-2}, \textbf{x}_{k, t-3}\}$ and $U_{i}$ is the noise affecting $\textbf{x}_{i, t}$. 

We independently model the local causal mechanisms for each effect variable $\widetilde{\textbf{x}}_{i, t}$ to obtain a causally disentangled representation.

\subsection{Casually Disentangled Representation Learning (CDRL)} \label{cdrl}

In this subsection, we explain how CDRL4AD captures causal relationships with causally disentangled representations, node and edge correlations, and temporal dependency representations. 

\subsubsection{\textbf{Causal Discovery}} \label{cd}
This component computes how cause variables influence on effect variables to determine if a causal relationship exists within the causal graph \( \mathcal{G}_{ca}(t) \) at time $t$.

Let $\ddot{\textbf{x}}_{j, t'}$ be a cause variable of $\widetilde{\textbf{x}}_{i, t}$, and let \( \alpha_{(i, t),(j, t')}^{CDR(t)} \) be an attention weight for the Causally Disentangled Representation (CDR), which signifies the importance of $\ddot{\textbf{x}}_{j, t'}$ in affecting $\widetilde{\textbf{x}}_{i, t}$. Our causal discovery process identifies candidate cause variables relevant to the effect variable and then selects the most significant cause variables from the candidates using the attention mechanism, denoted by $\mathcal{C}_i^{(t)}$:
\begin{equation}
    \begin{aligned}
    \alpha_{(i, t),(j, t')}^{CDR(t)} &= \frac{\exp(LeakyReLU(\textbf{a}^{T}[\widetilde{\textbf{x}}_{i, t} \| \ddot{\textbf{x}}_{j, t'}]))}{\sum_{k=1}^N \sum_{t-\omega \le t' < t} \exp(\text{LeakyReLU}(\textbf{a}^{T}[\widetilde{\textbf{x}}_{i, t} \| \ddot{\textbf{x}}_{k, t'}]))} \\ \\
    \mathcal{C}_i^{(t)} &= \{\ddot{\textbf{x}}_{j, t'} \mid \alpha_{(i, t),(j, t')}^{CDR(t)} \geq \theta\}
    \end{aligned}
\label{eq:causal_dis}
\end{equation}
where $\ddot{\textbf{x}}_{j, t'} \in \mathbb{R}^{m}$ and $\widetilde{\textbf{x}}_{i, t} \in \mathbb{R}^{m}$ denote the time series vectors from variables $j$ and $i$ at times $t'$ and $t$, respectively, with $t' < t$. The vector $\textbf{a}$ contains learned parameters used in the attention computation, and $\|$ denotes concatenation. A cause variable $\ddot{\textbf{x}}_{j, t'}$ is considered a true cause variable, and is included in the set $\mathcal{C}_i^{(t)}$, if and only if $\alpha_{(i, t),(j, t')}^{CDR(t)}$ exceeds or equals a predefined causal threshold $\theta$.

The candidate cause variables include all variables from time $t-\omega \leq t' < t$. Since all true cause variables are derived from past values and the direction of all edges in the causal graph points toward future times, the discovered SCM forms a Directed Acyclic Graph (DAG), illustrating how past data influences future outcomes.    

These time-lagged causal relationships are fed into the causally disentangled representation, as described in the following section.

\subsubsection{\textbf{Causally Disentangled Representation}}  
Our Causally Disentangled Representation (CDR) aligns latent dimensions with predefined causal relationships. Unlike existing DRL studies\cite{SS-FVAE-BB, CausalVAE, CaD-VAE}, we focus on disentangling causal factors relevant to anomalous behaviors in order to improve both detection accuracy and interpretability. The CDR delineates a sophisticated form of data representation, where the latent variables $Z^t$, produced by an encoder, encapsulate distinct and interpretable facets of the input data. Each facet directly corresponds to different causal mechanisms, as specified by a causal graph. A multi-head decoder utilizes these disentangled latent variables to reconstruct various features, with each feature being influenced by distinct causal pathways. 

To encode the causal relationships identified through causal discovery into the latent dimensions of the CDR, we introduce a causal relationship representation. Only cause variables from $\mathcal{C}_i^{(t)}$ are used to calculate the causal relationship representations, which are then passed to a multi-head variational autoencoder(VAE) to generate the CDR. Here, since VAEs are widely recognized for their ability to disentangle latent variables \cite{CausalVAE}, we incorporate VAEs in the multi-head decoder to achieve effective disentanglement.

The causal relationship representation $r_i^{(t)}\in\mathbb{R}^{l}$ between $\ddot{\textbf{x}}_{j, t'}$ and $\widetilde{\textbf{x}}_{i, t}$ is calculated by:
\begin{equation}
    r_i^{(t)} = \textit{LeakyReLU} \left( W_e S_i^{(t)} + \sum_{(j, t') \in \mathcal{C}_i^{(t)}} \alpha_{(i, t),(j, t')}^{CDR(t)} W_c \ddot{\textbf{x}}_{j, t'} \right)
\end{equation}
where $W_c \in \mathbb{R}^{l \times m}$ and $W_e \in \mathbb{R}^{l \times (\omega \times m)}$ are weighted matrices representing the cause and effect variables, respectively, and $l$ denotes the dimensionality of the embedding.

The causal relationship representation $r_i^{(t)}$ is then input into a multi-head decoder within VAE to generate the CDR, denoted as $d^{CDR(t)}$. In this VAE architecture, the encoder produces the mean $\mu$ and the standard deviation $\sigma$, and samples an epsilon $\epsilon$ to generate the latent space $Z^t$ using the reparameterization, assuming a Gaussian distribution. The latent variable $Z^t$ is passed through the decoder to reconstruct the input data. Unlike VAE, our model employs a multi-head decoder, which enhances the flexibility in generating multivariate outputs.

The encoder models the latent variable $Z^t\in\mathbb{R}^l$ by learning the posterior $q_{\phi}(Z^t \mid r^{(t)})$. The encoding process can be formalized as:
\begin{equation}
\begin{aligned}
    \tilde{h}^t &= \tanh(W_r r^t + W_h \tilde{h}^{t-1} + b) \\
    \mu &= W_{\mu} \tilde{h}^{t-1} + b_{\mu} \\
    \log(\sigma) &= W_{\sigma} \tilde{h}^{t-1} + b_{\sigma} \\
    \epsilon &= \mathcal{N}(0, I) \\
    Z^t &= \tanh(W_{re}(\mu + \sigma \epsilon) + b_{re})
\end{aligned}
\end{equation}
where $\tilde{h}^t\in\mathbb{R}^l$ denotes the hidden state of the encoder at time t, and $W_{\mu}$, $W_{\sigma}$,\(W_r\), \(W_h\),\(W_re\) are learned weight matrices corresponding to the respective variables and states. The terms $b_{\mu}$, $b_{\sigma}$, and $b_{re}$ are biases. The reparameterization of the latent variable $z^T$ makes the model differentiable, enabling the generation of high-quality reconstructions.

As shown in the CDR of Figure \ref{fig:1}, each decoder head disentangles the latent space $Z^t$ as a prior and independently generates the final hidden state \( \hat{h}_i^t \in \mathbb{R}^l \):
\begin{equation}
     \hat{h}_i^t = \tanh(W_{head(i)} S_i^{(t)} + W_{z} Z^t + b_{head(i)})
\end{equation}
where $head$($i$) function maps the index of the decoder head corresponding to the $i$-th variable and $W_{in(i)}$, $W_{z}$, and $b_i$ denote weights for the input and latent variable, and the bias of the $i$-th decoder, respectively. The $i$-th decoder head models the local causal mechanism \(f_i(PA(\widetilde{\textbf{x}}_{i, t}), U_i) \rightarrow \widetilde{\textbf{x}}_{i, t}\) into \( \hat{h}_i^t\) by estimating the posterior \( p_{\theta}(S_i^{(t)} | Z^t) \).

The causally disentangled representation $d^{CDR(t)} \in \mathbb{R}^{N \times l}$ is then formed by combining the $\hat{h}_i^t$ values from all variables:
\begin{equation}
    d^{CDR(t)} = \{\hat{h}_1^t, \hat{h}_2^t, \ldots, \hat{h}_N^t\}
\end{equation}

\subsubsection{\textbf{Node and Edge Correlation Representation}} 
The Node and Edge Correlation Representation (NECR) encodes variable correlations on two levels: within nodes and between nodes and edges. Our model learns the graph structure by connecting nodes that exhibit similar patterns, thus identifying neighboring nodes and edges based on the cosine similarity between node embeddings, which represent each node's characteristics. 

We initialize the node embeddings $b_i \in \mathbb{R}^l$ for each node $i \in \{1, 2, \ldots, N\}$ and these embeddings are learned during training. The cosine similarity $\mathcal{E}_{ij}$ between node $i$ and node $j$ is computed as:
\begin{equation}
\begin{aligned}
    \mathcal{E}_{ij} &= \frac{{b_i^T b_j}}{{\|b_i\|\|b_j\|}} \\
    \mathcal{N}_i^{(t)} &= \{j | j \in TopK(\{\mathcal{E}_{ij}\})\} 
\end{aligned}
\label{eq:neighbourK}
\end{equation}
where $\mathcal{E}_{ij}$ notes an edge feature, which can be enhanced with domain knowledge, and its dimensionality is denoted as $g$. Node $j$ is identified as a neighboring node of node $i$ and included in the set $\mathcal{N}_i^{(t)}$ if $\mathcal{E}_{ij}$ is among the top-$K$ similarities.  

The model then generates node-level representation $d^{n(t)}$ and edge-level representation $d^{e(t)}$ by aggregating node features and edge features, respectively:
\begin{equation}
\begin{aligned}
    d_i^{n(t)} &= LeakyReLU\left(\sum_{j \in \mathcal{N}_i^{(t)}} \alpha_{ij}^{NCR(t)} W_n S_i^{(t)}\right) \\
    d_i^{e(t)} &= LeakyReLU\left(\sum_{j \in \mathcal{N}_i^{(t)}} \alpha_{ij}^{ECR(t)} W_e \mathcal{E}_{ij}\right)
\end{aligned}
\end{equation}
where $\mathcal{E}_{ij}\in \mathbb{R}^{g}$ is the edge feature between node $i$ and $j$, $W_n \in \mathbb{R}^{l \times (\omega \times m)}$ and $W_e \in \mathbb{R}^{l \times g}$ are weighted matrices for linear transformation, and $\alpha_{ij}^{NCR(t)}$ and $\alpha_{ij}^{ECR(t)}$ are the node-level and edge-level attention weights, respectively.

The attention weights of node and edge features are computed as follows:
\begin{equation}
    \begin{aligned}
        \alpha_{ij}^{NCR(t)} &= \frac{\exp(\text{LeakyReLU}(\textbf{a}^{T}[W_n S_{i}^{(t)} \| W_n S_{j}^{(t)}]))}{\sum_{k \in \mathcal{N}_i^{(t)}}\exp(\text{LeakyReLU}(\textbf{a}^{T}[W_n S_{i}^{(t)} \| W_n S_{k}^{(t)}]))} \\
        \alpha_{ij}^{ECR(t)} &= \frac{\exp(\text{LeakyReLU}(\textbf{a}^{T}[W_n S_{i}^{(t)} \| W_e \mathcal{E}_{ij}]))}{\sum_{k \in \mathcal{N}_i^{(t)}}\exp(\text{LeakyReLU}(\textbf{a}^{T}[W_n S_{i}^{(t)} \| W_e \mathcal{E}_{ik}]))}
    \end{aligned}
\end{equation}
Here, $\textbf{a}$ is a parameter vector for attention computation, and $\|$ denotes concatenation.

The node-level and edge-level correlation representations are then concatenated to form the NECR $d^{NECR(t)} \in \mathbb{R}^{N \times 2l}$:
\begin{equation}
    d^{NECR(t)} = \{ d^{n(t)} \| d^{e(t)} \}
\end{equation}

\subsubsection{\textbf{Temporal Dependency Representation}}
We also learn the Temporal Dependency Representation (TDR) to capture the sequential relationship in data, where the current time step depends on previous time steps. Let \(j\) denote a past timestamp and let $i$ denote the present timestamp within the temporal dependency graph $\mathcal{G}_{td}^{(t)}$ such that $i \in \{t-\omega, \ldots, t-1\}$ and $j \in \{t-\omega, \ldots, t-1\} \backslash \{i\}$. 

The input vector of the TDR at time \(t\), as denoted as $ts^{(t)} = \{ts^{t-\omega(t)}, \ldots, ts^{t-1(t)}\}$, consists of observed vectors from the past $\omega$ time steps. Each time step contains observations from $\omega$ nodes, where each node $i$ is represented by a vector of observed values for $N$ variables at time $i$, expressed as $ts^{i(t)}=\{x_{1, i}, x_{2, i}, \ldots, x_{N, i}\}$. 

The temporal dependency representation \(d^{TDR(t)} \in \mathbb{R}^{N \times l}\) is computed by aggregating the input vector \(ts^{i(t)}\) of each node $i$ and its neighboring nodes \(ts^{j(t)}\):
\begin{equation}
    d_i^{TDR(t)} = LeakyReLU\left(\sum_{j=t-\omega}^{t-1} \alpha_{ij}^{TDR(t)} W ts^{j(t)}\right)
\end{equation}
where \(ts^{j(t)} \in \mathbb{R}^{N \times m}\) denotes the input vector for node $j$, and \(W \in \mathbb{R}^{l \times (N \times m)}\) is a learnable weight matrix.  

The attention weight $\alpha_{ij}^{TDR(t)}$ for the temporal dependency between different times is computed as follows:
\begin{equation}
    \begin{aligned}
        \alpha_{ij}^{TDR(t)} = \frac{\exp(LeakyReLU(a^{T}[Wts^{i(t)} \| Wts^{j(t)}]))}{\sum_{k=t-(w-1)}^{t} \exp(LeakyReLU(a^{T}[Wts^{i(t)} \| Wts^{k(t)}]))}
    \end{aligned}
\end{equation}

\subsubsection{\textbf{Output Layer}}
This layer generates \(d^{(t)} \in \mathbb{R}^{N \times l}\) by concatenating the causally disentangled, node and edge correlation, and temporal dependency representations:
\begin{equation}
    d^{(t)}={ GRU(d^{CDR(t)} \| d^{NECR(t)} \| d^{TDR(t)}) }
\end{equation}
The concatenated representations are fed into the gated recurrent unit (GRU) to convert them into a fixed-length output $d^{(t)}$. The output is then used to compute both the anomaly score and the root cause score (as described in Section \ref{ad} and \ref{rce}). Specifically, $d^{(t)}$ is used to predict $x_{i, t}$ and reconstruct ${S}_i^t$ through fully connected layers and a VAE, respectively.

\subsection{Objective Function and Training} \label{ad}
Our objective function is jointly optimized by incorporating both prediction loss and reconstruction loss in order to achieve accurate anomaly detection. Given a specific time \( t \), the objective function \( \mathcal{L}^{(t)} \) is defined as:
\begin{equation}
    \mathcal{L}^{(t)} = \mathcal{L}_{f}^{(t)} + \mathcal{L}_{r}^{(t)}
\end{equation}
 where \( \mathcal{L}_{f}^{(t)} \) denotes the prediction loss, and $\mathcal{L}_{r}^{(t)}$ denotes the reconstruction loss.

It is worth noting that the joint optimization of both losses allows the model to compute the anomaly score effectively, particularly during anomaly occurrence. This enables the model to detect anomalies by identifying shifts in data distribution and patterns.

\textbf{Prediction Loss}: quantifies the deviation between the predicted and true future values. Our model improves prediction accuracy by disentangling causal relationships, which allows it to capture changes in relationships between individual nodes.
The prediction loss, denoted as $\mathcal{L}_{f}^{t}$, is calculated using mean squared error.
\begin{equation}
    \mathcal{L}_{f}^{(t)} = (X^t - \hat{X}^t)^2 
\end{equation}
where $X^t$ denotes the true time series vector at time $t$, and $\hat{X}^t$ is its prediction.

\textbf{Reconstruction Loss}: quantifies the difference in probability distribution between the input and output of the model. We reconstruct \( S^{(t)} \) into $\hat{S}^{(t)}$ through a VAE. With encoder $q$ and decoder $p$, the reconstruction loss $\mathcal{L}_{r}^{(t)}$ is defined as:
 \begin{equation}
    \mathcal{L}_{r}^{(t)} = - E_{q_{\phi}(Z^t|d^{(t)})}[logp_{\theta}(S^{(t)}|Z^t)] + D_{KL}(q_{\phi}(Z^t|d^{(t)}) \| p_{\theta}(Z^t))
\end{equation}
where $D_{KL}$ refers to the Kullback-Leibler(KL) divergence, $S^{(t)}$ denotes the set of samples at time $t$, and $d^{(t)}$ denotes the output of the CDRL.

The first term describes the difference between the input $S^{(t)}$ and its reconstruction $\hat{S}^{(t)}$ in terms of probability distribution. The second term, $D_{KL}$, serves as regularization to guide the encoder $q$ in learning latent variables that approximate the Gaussian distribution aligned with the prior distribution $p_{\theta}(Z^t)$. The combination of the reconstruction probability and KL divergence yields the total reconstruction loss $\mathcal{L}_{r}^{(t)}$.

\subsection{Anomaly and Root Cause Examination} \label{rce}
An anomaly score reflects the degree of anomalousness at time $t$. When an anomaly occurs, each variable at time $t$ has a root cause score that represents the anomalousness for that specific variable. The root cause scores enable effective diagnosis by identifying the underlying issues causing the anomaly. This facilitates efficient problem-solving as it provides clear insights into which parts of the system are anomalous.

We formally define the anomaly score function $\mathcal{A}(t)$ as:
\begin{equation}
\begin{aligned}
    \mathcal{A}(t) = \frac{1}{N} \sum_{i=1}^N rs_i(t)
\end{aligned}
\end{equation}

where the function $rs_i(t)$ computes the root cause score for variable $i$ at time $t$ by combining both prediction and reconstruction errors. 

In detail, the function $rs_i(t)$ is also defined as the follows:
\begin{equation}
    rs_i(t) = \sqrt{(x_{i, t} - \hat{x}_{i, t})^2 + \beta({S}_i^{(t)} - \tilde{S}_i^{(t)})^2}
    \label{eq:rmse}
\end{equation}
where $\beta$ is a hyperparameter used to merge the two errors based on data characteristics and ${S}_i^{(t)} \in \mathbb{R}^{\omega \times m}$ denotes the sample set of variable $i$ at time $t$. 

Here, the root mean squared error (RMSE) is used to combine these errors based on the forecasted observation value $\hat{x}_{i, t}$ and the reconstructed sample set $\tilde{S}_i^{(t)}$.

The model diagnoses the time $t$ as an anomaly if $\mathcal{A}(t)$ exceeds a predefined threshold, automatically determined using the Peak Over Threshold (POT) \cite{POT(Peak-Over-Threshold)}, as illustrated in the right part of Figure \ref{fig:1}. Then the model localizes the variables with Top-$k$ root cause scores among $R(t) = \{rs_1(t), \ldots, rs_N(t)\}$ as root causes at the anomalous time $t$.

\section{Evaluation} \label{evaluation}
In this section, we conducted a comparative analysis of our model with state-of-the-art methods in terms of accuracy and root cause analysis. We also carried out an ablation study to validate the contribution of our components. We measured the in-depth analysis of our model on sensitivity and performance overhead.
The code repository is publicly available and also accessible via \textit{\url{https://github.com/datascience-labs/CDRL4AD}}.

\subsection{Experimental Design}
The datasets, baselines and evaluation metrics, research questions, and experimental setup were presented.

\subsubsection{\textbf{Datasets}}
Table \ref{tab:datasets} describes the statistics of the eight datasets used for evaluation: the number of features, training set size, testing set size, and anomaly rates (AR). 
\begin{itemize}
    \item Secure Water Treatment (\textbf{SWaT}) included attacks simulated on a water treatment plant testbed created by the iTrust Centre \cite{SWaT}.
    \item The Server Machine Dataset (\textbf{SMD}) were collected from a large Internet company on 28 server machines and contains root cause labels provided by domain experts for test sets. We extracted subsets from machine-1-6, 2-8, and 3-6 in SMD\cite{OmniAnomaly} . 
    \item The \textbf{HAI} dataset were collected from a Hardware-In-the-Loop testbed that integrates water treatment, boiler, and turbine processes, by the National Security Research Institute\cite{HAI}.
    \item The Pooled Server Metrics (\textbf{PSM}) dataset consisted of 25 server machine metrics from eBay\cite{PSM}. 
    \item The Soil Moisture Active Passive (\textbf{SMAP}) dataset contained soil samples and continuous measurement data collected by NASA's Mars exploration rover \cite{SMAP_MSL}.
    \item The Mars Science Laboratory (\textbf{MSL}) contained both sensor and actuator data from the Mars exploration spacecraft itself \cite{SMAP_MSL}.
    \item The Water Distribution (\textbf{WADI}), an extension of the SWaT dataset, included a larger number of sensors and a more realistic water treatment system, but it is highly noisy, as mentioned in \cite{TranAD}, \cite{WADI}.
    \item The Global Content Platform (\textbf{GCP}) dataset contained performance monitoring metrics sampled every five minutes from 30 online service systems over seven weeks \cite{GCP}.
\end{itemize}

Note that these datasets exhibited time-lagged causal relationships and correlations among variables under MTS. 

\begin{table}[b]
\centering  \caption{Datasets statistics} \label{tab:datasets}
\begin{tabular}{crrrr}
\hline
\textbf{Datasets}     & \textbf{\#Features} & \textbf{Train. size} & \textbf{Test. size} & \textbf{AR(\%)} \\ \hline
SWaT\cite{SWaT}       & 51                  & 49,680                      & 44,991                     & 12.09                     \\
HAI\cite{HAI}         & 59                  & 55,080                      & 44,460                     & 3.94                      \\
SMD\cite{OmniAnomaly} & 38                  & 76,117                      & 81,158                     & 3.33                      \\
PSM\cite{PSM}         & 25                  & 129,784                     & 87,841                     & 27.76                     \\ 
SMAP\cite{SMAP_MSL}                  & 25                  & 135,183                     & 427,617                    & 13.13                     \\
MSL\cite{SMAP_MSL}                   & 55                  & 58,317                      & 73,729                     & 10.72                     \\
WADI\cite{WADI}                  & 127                 & 78,457                      & 17,280                     & 5.99                      \\ 
GCP\cite{JumpStarter}           & 19                  & 172,800                     & 216,000                    & 20.26      \\ \hline
\end{tabular}
\end{table}

\subsubsection{\textbf{Baselines and evaluation metrics}} 
We assessed the performance of \textbf{CDRL4AD}, compared to state-of-the-art methods including GNN-based methods: \textbf{MTAD-GAT} \cite{MTAD-GAT}, \textbf{GDN} \cite{GDN}, \textbf{DuoGAT} \cite{DuoGAT}, and \textbf{FuSAGNet} \cite{FuSAGNet}, as well as Deep Learning(DL)-based methods: \textbf{DAGMM} \cite{DAGMM}, \textbf{USAD} \cite{USAD}, \textbf{MAD-GAN} \cite{MAD-GAN}, and \textbf{TranAD} \cite{TranAD}. We excluded models based on CRL or DRL because they have not been studied for anomaly detection.

We used common metrics to facilitate comparative analysis with state-of-the-art methods for detection accuracy: Precision, Recall, F1-Score, and AUC. The point-adjustment strategy from \cite{point-adjust} was used to determine anomaly detection performance. We thus regarded an anomaly as correctly identified if any sample within the anomaly sector is detected as an anomaly state.

To evaluate the effectiveness of root cause analysis, we adopted both HitRate@P\% (H) \cite{OmniAnomaly} and Normalized Discounted Cumulative Gain (NDCG)@P\% (N) \cite{ndcg} metrics. Here, $P\%$ denotes the percentage of top-predicted root causes relative to the number of ground truth root causes. For instance, the top 15 root cause candidates are retrieved by the model when P=150 and 10 ground truth root causes exist in a certain anomaly. HitRate measures the proportion of correctly predicted root cause candidates that match the ground truth without considering their ranking. In contrast, NDCG gives higher scores to the correctly predicted root cause candidates that are ranked higher to measure the ranking quality.

For all six metrics, `1' represents the best performance, while `0' represents the worst.

\subsubsection{\textbf{Research questions}}
To evaluate our approach, we aimed at answering the following research questions:
\begin{itemize}
    \item (\verb|ACCURACY|) How effective is our model in detecting anomalies compared to state-of-the-arts methods? 
    \item (\verb|ABLATION STUDY|) How much does each component of our model affect detecting anomalies?
    \item (\verb|ROOT CAUSE ANALYSIS|) Does our model effectively identify the root causes of anomalies?
    \item (\verb|SENSITIVITY|) How sensitive is our model to variations in hyper-parameters?
    \item (\verb|TIME COMPLEXITY|) How efficient is our model in practical scenarios in terms of time complexities? 
\end{itemize}

\subsubsection{\textbf{Experimental setup}}
Both our model and baselines were implemented in PyTorch 2.0 and supported by CUDA 11.8, on a server equipped with an Intel Xeon Platinum 8468 48C CPU at 2.10$GHz$ and dual RTX A6000 GPUs (48GB VRAM each). All models were trained using Adam optimizer of the initial learning rate $1 \times 10^{-3}$ with 32 sized mini-batch. To prevent overfitting, an early stopping mechanism with 10 epoch tolerance was applied. In our model, the width \( \omega \) of the sliding window was set to $100$ with embedding dimension $l=64$. For NECR, $K=20$ was used.

\begin{table*}[t]
\caption{Anomaly detection performance}
\label{tab:accuracy}
\centering
\resizebox{1.0\textwidth}{!} {
\begin{tabular}{crrrrrrrrrrrrrrrr}               
\hline
\multirow{2}{*}{Method} & \multicolumn{4}{c}{SWaT}                                              & \multicolumn{4}{c}{HAI}                                                     & \multicolumn{4}{c}{SMD}                                               & \multicolumn{4}{c}{PSM}                                               \\ \cline{2-17} 
                 & Precision             & Recall             & F1              & AUC             & Precision                   & Recall             & F1              & AUC             & Precision             & Recall             & F1              & AUC             & Precision             & Recall             & F1              & AUC             \\ \hline
DAGMM\cite{DAGMM}                   & 0.2387          & 0.4721          & 0.3171          & 0.5122          & 0.1492                & 0.2920           & 0.1975          & 0.4997          & 0.1942          & 0.6350           & 0.2975          & 0.5011          & 0.3754          & 0.5054          & 0.4300          & 0.4899          \\
USAD\cite{USAD}                    & 0.7962          & 0.5765          & 0.6688          & 0.6968          & 0.3225                & \textbf{0.5755}          & 0.4133          & 0.7268          & 0.3544          & 0.5978          & 0.4450           & 0.6489          & 0.4203          & 0.7758          & 0.5444          & 0.5647          \\
MAD-GAN\cite{MAD-GAN}                 & 0.6082          & \textbf{0.6236} & 0.6149          & 0.6786          & 0.5055                & 0.4381          & 0.4694          & 0.7232          & 0.6012          & 0.5987          & 0.5999          & 0.6967          & 0.4237          & 0.8799          & 0.5718          & 0.5727          \\
TranAD\cite{TranAD}                  & 0.6878          & 0.5821          & 0.6306          & 0.6732          & 0.6899                & 0.4164          & 0.5193          & {\underline{0.7695}}    & 0.7112          & 0.5987          & 0.6501          & 0.7052          & 0.4446          & 0.8802          & 0.5909          & 0.5791          \\
GDN\cite{GDN}                     & 0.9843          & 0.5904          & 0.7381          & {\underline{0.7991}}    & 0.5755                & 0.4740           & 0.5197          & 0.7393          & 0.7699          & \textbf{0.6654} & 0.7138          & 0.8639          & {\underline{0.4639}}    & {\underline{0.9138}}    & {\underline{0.6152}}    & \textbf{0.7619} \\
MTAD-GAT\cite{MTAD-GAT}                & {\underline{0.9953}}    & 0.5878          & {\underline{0.7391}}    & 0.7969          & {\underline{0.8555}} & 0.3840           & 0.5300            & 0.6991          & \textbf{0.9548} & 0.6251          & {\underline{0.7555}}    & 0.7931          & 0.4636          & 0.9047          & 0.6128          & 0.7284          \\
FuSAGNet\cite{FuSAGNet}                & 0.9878          & 0.5860           & 0.7356          & 0.7692          & 0.7681                & 0.4163          & 0.5399          & 0.7439          & 0.7711          & {\underline{0.6598}}    & 0.7111          & {\underline{0.8971}}    & 0.4611          & 0.8645          & 0.6009          & 0.7118          \\
DuoGAT\cite{DuoGAT}                  & 0.6986          & {\underline{0.6235}}    & 0.6589          & 0.7113          & 0.8024                & 0.5199          & {\underline{0.6309}}    & \textbf{0.7972} & 0.7204          & 0.6075          & 0.6592          & 0.8248          & 0.4548          & 0.8667          & 0.5968          & 0.6550          \\ \hline 
\textbf{CDRL4AD}        & \textbf{0.9970} & 0.6053          & \textbf{0.7526} & \textbf{0.8060} & \textbf{0.9884}       & {\underline{0.5569}}    & \textbf{0.7117} & 0.7598          & {\underline{0.9482}}    & 0.6435          & \textbf{0.7667} & \textbf{0.9261} & \textbf{0.4859} & \textbf{0.9207} & \textbf{0.6361} & {\underline{0.7572}}    \\ \hline \hline
\multirow{2}{*}{Method} & \multicolumn{4}{c}{SMAP}                                              & \multicolumn{4}{c}{MSL}                                                     & \multicolumn{4}{c}{WADI}                                              & \multicolumn{4}{c}{GCP}                                       \\ \cline{2-17} 
     & Precision             & Recall             & F1              & AUC             & Precision                   & Recall             & F1              & AUC             & Precision             & Recall             & F1              & AUC             & Precision             & Recall             & F1              & AUC             \\ \hline 
DAGMM\cite{DAGMM}                   & 0.2201          & 0.2772          & 0.2454          & 0.5329          & 0.1537                & 0.2458          & 0.1892          & 0.5384          & 0.1069          & 0.2096          & 0.1395          & 0.5827          & 0.2795          & 0.2617          & 0.2703          & 0.5453          \\
USAD\cite{USAD}                    & 0.1549          & \textbf{0.9695} & 0.2671          & 0.3938          & 0.2403                & 0.5335          & 0.3314          & 0.5998          & 0.1651          & 0.875           & 0.2778          & 0.5557          & 0.2923          & \textbf{0.8450} & 0.4344          & 0.5385          \\
MAD-GAN\cite{MAD-GAN}                 & 0.8518          & 0.5663          & 0.6803          & 0.7759          & 0.8364                & 0.6402          & 0.7253          & 0.8127          & 0.0825          & 0.2642          & 0.1460          & 0.4625          & 0.2563          & 0.4305          & 0.3213          & 0.5044          \\
TranAD\cite{TranAD}                  & 0.9655          & {\underline{0.6757}}    & \textbf{0.7950} & \textbf{0.8359} & 0.5263                & \textbf{0.9999} & 0.6896          & \textbf{0.9975} & 0.1198          & \textbf{0.9999} & 0.2140          & 0.6305          & 0.3788          & 0.3829          & 0.3809          & 0.6420          \\
GDN\cite{GDN}                     & {\underline{0.9810}}    & 0.5360          & 0.6932          & 0.7672          & 0.8987                & 0.5989          & 0.7188          & 0.7955          & {\underline{0.3126}}    & 0.9293          & 0.2662          & 0.6320          & 0.9768          & 0.1502          & 0.2604          & 0.5746          \\
MTAD-GAT\cite{MTAD-GAT}                & \textbf{0.9859} & 0.5465          & 0.7032          & 0.7726          & \textbf{0.9915}       & 0.6132          & 0.7577          & 0.8063          & 0.0890          & 0.9327          & 0.1625          & {\underline{0.6722}}    & 0.6957          & 0.6091          & {\underline{0.6495}}    & {\underline{0.7707}}    \\
FuSAGNet\cite{FuSAGNet}                & 0.9615          & 0.5575          & 0.7058          & 0.7771          & 0.9181                & 0.7432          & 0.8215          & 0.8677          & 0.0850           & {\underline{0.8736}}    & 0.1474          & 0.6312          & {\underline{0.9747}}    & 0.1348          & 0.3392          & 0.5670          \\
DuoGAT\cite{DuoGAT}                  & 0.9387          & 0.5578          & 0.6998          & 0.7762          & 0.9124                & {\underline{0.8542}}    & \textbf{0.8824} & {\underline{0.9223}}    & 0.2688          & 0.4042          & {\underline{0.3229}}    & 0.6682          & \textbf{0.9975} & 0.2783          & 0.4352          & 0.6391          \\ \hline
\textbf{CDRL4AD}        & 0.9665          & 0.5637          & {\underline{0.7121}}    & {\underline{0.7804}}    & {\underline{0.9509}}          & 0.7613          & {\underline{0.8456}}    & 0.9106          & \textbf{0.7303} & 0.3640          & \textbf{0.4859} & \textbf{0.6779} & 0.8749          & {\underline{0.7739}}    & \textbf{0.8213} & \textbf{0.8729} \\ \hline
\end{tabular}
}
\begin{flushleft}
\footnotesize{* The top results for each metric are shown in boldface and the second-best results are
underlined.}\\
\end{flushleft}
\end{table*}

\subsection{Accuracy}
Table \ref{tab:accuracy} presents the results of the anomaly detection performance evaluation. CDRL4AD demonstrated superior performance in terms of F1 and AUC scores across all datasets. Notably, our model recorded the highest F1 score in all datasets except SMAP and MSL while consistently ranking at least third in AUC across all datasets. This result demonstrates its balanced and reliable performance.

GNN-based methods outperformed DL-based approaches across most MTS datasets. The finding highlights the necessity of capturing both edge correlations and dynamic node relationships. However, most GNN-based methods exhibited reduced performance in F1 scores, compared to our model. The performance degradation was due to difficulties in capturing complex relationships in MTS contexts. Specifically, both GDN and FuSAGNet were limited in fully reflecting the dynamic temporal aspects of feature correlations because they were built upon static graph structures. MTAD-GAT disregarded changes in both structural and feature relationships across most datasets due to its underlying assumption of fully connected graphs. Meanwhile, DuoGAT had a hardship to capture complex patterns due to its heavy reliance on the difference of time series.

CDRL4AD achieved the highest F1 and AUC scores even on noisy datasets like WADI, compared the state-of-the-art methods. They failed to infer directional causality since they were restricted to undirected graph structure. In contrast, CDRL4AD explicitly identified time-lagged causal relationships through causal discovery, simultaneously modeling correlation and temporal dependencies.

\begin{table*}[t]
\caption{Impact of component removal on model performance}
\label{tab:ablation}
\centering
\resizebox{1.0\columnwidth}{!}{%
\begin{tabular}{ccrrrrrrrrrrrrrrrr}               
\hline
\multirow{2}{*}{Category} & \multirow{2}{*}{Configuration} & \multicolumn{4}{c}{SWaT} & \multicolumn{4}{c}{HAI} & \multicolumn{4}{c}{SMD} & \multicolumn{4}{c}{PSM} \\ \cline{3-18} 
 &  & Precision & Recall & F1 & AUC & Precision & Recall & F1 & AUC & Precision & Recall & F1 & AUC & Precision & Recall & F1 & AUC \\ \hline
Full model & \textbf{CDRL4AD (Ours)} & 0.9970 & 0.6053 & \textbf{0.7526} & \textbf{0.8060} & 0.9884 & \textbf{0.5569} & \textbf{0.7117} & \textbf{0.7598} & 0.9482 & \textbf{0.6435} & \textbf{0.7667} & \textbf{0.9261} & \textbf{0.4859} & 0.9207 & \textbf{0.6361} & \textbf{0.7572} \\ \hline
Individual & w/o TDR & 0.9787 & 0.5989 & 0.7430 & 0.7923 & 0.9869 & 0.3436 & 0.5097 & 0.6791 & 0.9594 & 0.6342 & 0.7636 & 0.9185 & 0.4740 & 0.9259 & 0.6270 & 0.7050 \\
component & w/o Edge & 0.9834 & 0.5976 & 0.7433 & 0.8038 & 0.9880 & 0.3265 & 0.4908 & 0.7061 & \textbf{0.9491} & 0.6368 & 0.7622 & 0.8967 & 0.4605 & 0.9051 & 0.6104 & 0.7004 \\
removals & w/o NECR & \textbf{0.9975} & 0.5882 & \textbf{0.7526} & 0.7920 & 0.7526 & 0.3686 & 0.4948 & 0.6735 & 0.9261 & 0.6160 & 0.7399 & 0.9254 & 0.4305 & 0.9139 & 0.5853 & 0.6901 \\
 & w/o DRL & 0.9512 & 0.6125 & 0.7451 & 0.7981 & 0.9738 & 0.3595 & 0.5251 & 0.7156 & 0.9428 & 0.6378 & 0.7608 & 0.8835 & 0.4356 & \textbf{0.9286} & 0.6157 & 0.6950\\
 & w/o CRL & 0.9510 & 0.6031 & 0.7381 & 0.7966 & \textbf{0.9895} & 0.3214 & 0.4852 & 0.6740 & 0.9354 & 0.6354 & 0.7568 & 0.8746 & 0.4356 & 0.7994 & 0.5639 & 0.6805 \\ \hline
Combined & w/o CRL\&DRL(CDR) & 0.9555 & 0.5922 & 0.7312 & 0.7797 & 0.9894 & 0.3174 & 0.4806 & 0.6433 & 0.8520 & 0.6231 & 0.7198 & 0.8834 & 0.4123 & 0.8043 & 0.5451 & 0.6568 \\
component & w/o NECR\&TDR & 0.9549 & \textbf{0.6158} & 0.7487 & 0.7875 & 0.6502 & 0.3965 & 0.4926 & 0.6938 & 0.6857 & 0.7234 & 0.7007 & 0.8529 & 0.4075 & 0.8313 & 0.5469 & 0.6754 \\
removals & w/o NECR\&CDR & 0.9590 & 0.5768 & 0.7203 & 0.7867 & 0.8865 & 0.3279 & 0.4787 & 0.6631 & 0.6669 & 0.6921 & 0.6740 & 0.8362 & 0.3787 & 0.7929 & 0.5126 & 0.6341 \\
 & w/o TDR\&CDR & 0.9615 & 0.5884 & 0.7300 & 0.7839 & 0.7799 & 0.3176 & 0.4514 & 0.6570 & 0.6630 & 0.7956 & 0.7104 & 0.8853 & 0.4041 & 0.8257 & 0.5426 & 0.6704 \\ \hline \hline
\multirow{2}{*}{Category} & \multirow{2}{*}{Configuration} & \multicolumn{4}{c}{SMAP} & \multicolumn{4}{c}{MSL} & \multicolumn{4}{c}{WADI} & \multicolumn{4}{c}{GCP} \\ \cline{3-18} 
 &  & Precision & Recall & F1 & AUC & Precision & Recall & F1 & AUC & Precision & Recall & F1 & AUC & Precision & Recall & F1 & AUC \\ \hline
Individual & w/o TDR & 0.9651 & 0.5586 & 0.7076 & 0.7778 & 0.8657 & 0.7593 & 0.8090 & 0.8727 & 0.7303 & 0.3640 & 0.4859 & 0.6779 & 0.8644 & 0.6813 & 0.7620 & 0.8270 \\
component & w/o Edge & 0.9771 & 0.5532 & 0.7064 & 0.7766 & 0.9152 & 0.7390 & 0.8177 & 0.8655 & 0.5874 & 0.3641 & 0.4495 & 0.6742 & 0.8976 & 0.6620 & 0.7620 & 0.8214 \\
removals & w/o NECR & \textbf{0.9881} & 0.5480 & 0.7050 & 0.7784 & 0.9013 & 0.6964 & 0.7857 & 0.8726 & 0.6696 & 0.3009 & 0.4152 & 0.6458 & 0.9073 & 0.6289 & 0.7429 & 0.8063 \\
 & w/o DRL & 0.9505 & 0.5602 & 0.7049 & 0.7780 & 0.7871 & 0.7547 & 0.7706 & 0.8653 & 0.5874 & 0.3641 & 0.4495 & 0.6742 & 0.9022 & 0.6515 & 0.7566 & 0.8168 \\
 & w/o CRL & 0.9548 & 0.5583 & 0.7046 & 0.7772 & 0.9136 & 0.6768 & 0.7776 & 0.8346 & 0.5902 & 0.3641 & 0.4504 & 0.6743 & \textbf{0.9547} & 0.6355 & 0.7630 & 0.8139 \\ \hline
Combined & w/o CRL\&DRL(CDR) & 0.9456 & 0.5616 & 0.7047 & 0.7735 & 0.7218 & 0.7990 & 0.7584 & 0.8355 & 0.6452 & 0.3009 & 0.4104 & 0.6454 & 0.9104 & 0.6209 & 0.7383 & 0.8027 \\
component & w/o NECR\&TDR & 0.9579 & 0.5567 & 0.7042 & 0.7756 & 0.7121 & 0.8291 & 0.7662 & 0.8948 & 0.9231 & 0.3009 & 0.4539 & 0.6497 & 0.8212 & 0.6619 & 0.7330 & 0.8126 \\
removals & w/o NECR\&CDR & 0.9727 & 0.5509 & 0.7034 & 0.7743 & 0.7928 & 0.7073 & 0.7476 & 0.8428 & 0.4184 & \textbf{0.4919} & 0.3641 & 0.6705 & 0.8177 & 0.6157 & 0.7022 & 0.7902 \\
 & w/o TDR\&CDR & 0.9264 & 0.5653 & 0.7022 & 0.7756 & 0.9579 & 0.6209 & 0.7534 & 0.8088 & 0.4294 & 0.3932 & 0.4104 & 0.6805 & 0.7989 & 0.6051 &  0.6886 & 0.7832 \\ \hline
\end{tabular}
}
\begin{flushleft}
\footnotesize{* The top results for each metric are shown in boldface.}\\
\end{flushleft}
\end{table*}
\vspace{-0.3cm}

\subsection{Ablation study}
In this study, we measured the impact of removing individual and combined components: TDR, NECR, and CDR. Specifically, we evaluated configurations of five components: replacing TDR's attention mechanism with GRU (\textit{w/o TDR}), removing edge-correlation features (\textit{w/o Edge}), substituting node and edge attention with GCN (\textit{w/o NECR}), replacing the multi-head decoder with a single-head VAE decoder (\textit{w/o DRL}), and substituting causal discovery components with an encoder structure (\textit{w/o CRL}). 

For combined component removals, we assessed the following: CDR (\textit{w/o CRL\&DRL}), NECR and TDR (\textit{w/o NECR\&TDR}), NECR and CDR (\textit{w/o NECR\&CDR}), and TDR and CDR (\textit{w/o TDR\&CDR}). Besides, we analyzed combinations by simultaneously removing CRL and DRL (\textit{w/o CRL\&DRL (CDR)}), TDR and NECR (\textit{w/o NECR\&TDR}), NECR and CDR (\textit{w/o NECR\&CDR}), and TDR and CDR (\textit{w/o TDR\&CDR}).

Table \ref{tab:ablation} shows the impact of the component removal on model performance. In general, the removal of the individual components resulted in better performance compared to the removal of the combined components, as measured by both both F1 and AUC scores. 

The results of the individual component removals revealed that the largest performance drop in F1 and AUC occurred in \textit{w/o CRL}. The degradation indicates that learning causal relationship learning in MTS contexts plays a pivotal role in capturing the varying interventions caused by past events. The second and third largest performance declines were observed in \textit{w/o NECR} and \textit{w/o DRL}, respectively, while the difference between them is trivial. The marginal difference implies that learning correlations between individual variables and disentanglement of latent variables play equally important roles. Although the performance drop in F1 and AUC in \textit{w/o TDR} was relatively minor, the reduction remained s substantial when compared to the full model.

In the results of combined component removals, \textit{w/o NECR\&CDR} exhibited the most significant drops in F1 and AUC, followed by \textit{w/o TDR\&CDR}, \textit{w/o CRL\&DRL (CDR)}, and \textit{w/o NECR\&TDR}. Especiall, \textit{w/o NECR\&TDR} achieved a minimum loss in accuracy, while others underwent substantial declines. For instance, the F1 and AUC scores in \textit{w/o NECR\&CDR} were dropped up to 23.30\% and 12.31\%, respectively, compared to the full model. 

\vspace{-0.3cm}
\begin{table}[b]
\caption{Comparison of root cause analysis methods}
\centering  \footnotesize 
\label{tab:rca}
\begin{tabular}{ccccc}
\hline
\multirow{2}{*}{Method} & \multicolumn{4}{c}{SMD} \\ \cline{2-5} 
 & H@100\% & H@150\% & N@100\% & N@150\% \\ \hline
DAGMM\cite{DAGMM} & 0.5825 & 0.6980 & 0.5853 & 0.6524 \\
USAD\cite{USAD} & 0.7681 & 0.8756 & 0.7960 & 0.8611 \\
MAD-GAN\cite{MAD-GAN} & 0.6187 & 0.7482 & 0.5977 & 0.6760 \\
TranAD\cite{TranAD} & 0.5766 & 0.7649 & 0.5968 & 0.7115 \\
GDN\cite{GDN} & 0.5635 & 0.7900 & 0.6734 & 0.7711 \\
MTAD-GAT\cite{MTAD-GAT} & 0.7386 & 0.8602 & 0.7733 & 0.8479 \\
FuSAGNet\cite{FuSAGNet} & 0.6403 & 0.7698 & 0.6748 & 0.7540 \\
DuoGAT\cite{DuoGAT} & 0.6604 & 0.7997 & 0.6979 & 0.7829 \\ \hline
CDRL4AD & \textbf{0.7719} & \textbf{0.8972} & \textbf{0.8057} & \textbf{0.8823} \\ \hline
\addlinespace[3pt]
\multicolumn{5}{l}{\footnotesize * The top results for each metric are shown in boldface.} \\
\end{tabular}
\end{table}

\subsection{Root Cause Analysis}
Root cause analysis (RCA) aimed to localize the root cause variables responsible for a detected anomaly at a specific time. CDRL4AD calculated root cause scores for variables causing anomalies using Eq. \eqref{eq:rmse}. Then, the Top-$k$ variables with the highest scores were retrieved as the root cause candidates for the detected anomaly at a specific time. Here, RCA was conducted only on the SMD dataset because other datasets lack ground truth labels for root causes.

Table \ref{tab:rca} presents the comparative analysis of our model and state-of-the-art methods in terms of HitRate (H) and NDCG (N). Specially, CDRL4AD outperformed state-of-the-art methods across all metrics, with HitRate of 0.7719 (H@100\%) and 0.8972 (H@150\%) and NDCG of 0.8057 (N@100\%) and 0.8823 (N@150\%). The results demonstrated that our model guarantees better identification of root causes of anomalies as well as their improved ranking quality.

Compared to USAD as the next-best method, CDRL4AD achieved the improvement of 2.16\% in H@150\% and 2.12\% in N@150\%. Other methods such as DAGMM and GDN failed to obtain high ranking quality in N@100\%. The reason for lower ranking quality was because their anomaly diagnosis lacks to understand causal relationships.

DL-based (DAGMM, USAD, MAD-GAN, and TranAD) and GNN-based methods (GDN, MTAD-GAT, FuSAGNet, and DuoGAT) exhibited similar performance in terms of HitRate. However, GNN-based methods obtained higher NDCG scores, which indicates better ranking quality. Again, CDRL4AD attained significantly higher precision in identifying root causes and improving ranking quality. The reason for the improvement was that our model captures causal relationships through causal discovery.

\begin{figure}[t]
\centering
\includegraphics[width=\columnwidth]{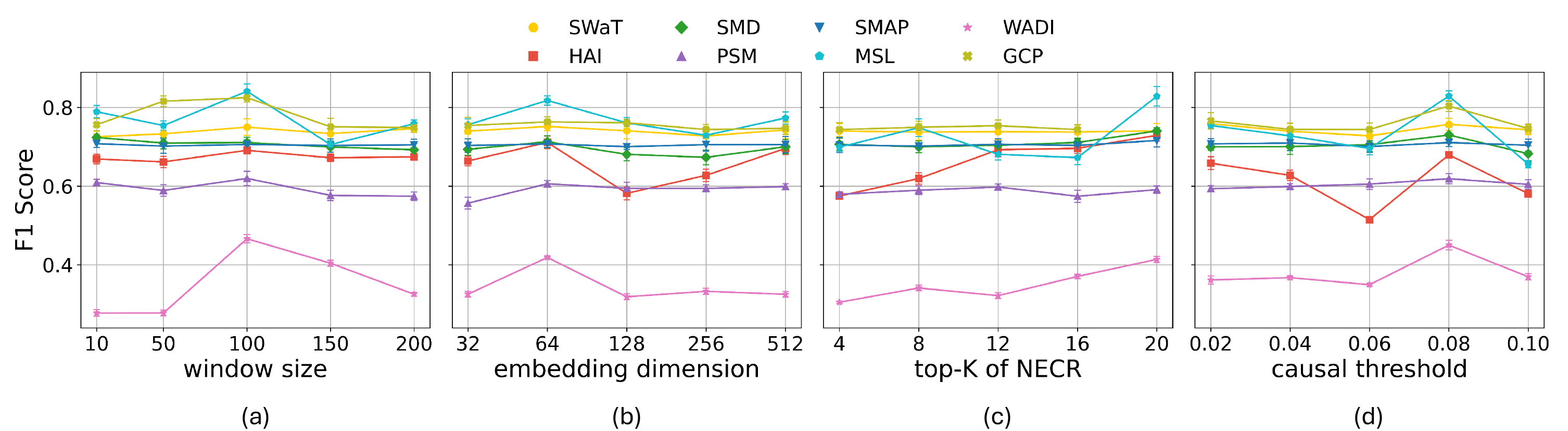}
\captionsetup{justification=centering}
\caption{Sensitivity Analysis}
\label{fig:Parameter Analysis} 
\end{figure}

\subsection{Sensitivity Analysis}
To understand the impact of key hyper-parameters on model performance, we conducted a sensitivity analysis on four parameters: sliding window size \(\omega\), embedding dimension \(l\), \(K\) for the selection of the top-\(K\) neighbor in Eq. \eqref{eq:neighbourK} from NECR component, and causal threshold $\theta$ in Eq. \eqref{eq:causal_dis}.

Figure \ref{fig:Parameter Analysis} (a)-(d) shows the average F1 scores of CDRL4AD with regard to the variations in the four hyper-parameters. Especially, Figure \ref{fig:Parameter Analysis} (a) shows the effect of window size \(\omega\). In the case of $\omega$=100, CDRL4AD exhibited the highest performance. Stable performance was observed with slight improvement within the rang of 50 to 150. 
Figure \ref{fig:Parameter Analysis} (b) shows the impact of changes in the embedding dimension \(l\) (32 to 512) on the F1 score. Our model achieved the best performance when $l$=64. In particular, F1 scores remain relatively stable across all datasets, with only minor variations as the embedding dimension increases. Figure \ref{fig:Parameter Analysis} (c) shows the impact of changes in top-$K$ value of NECR on the F1 score. The results exhibited the highest F1 score when $K$ = 20. Meanwhile, for the GCP dataset, $K$ was measured up to 16 because the number of its nodes is 19. Figure \ref{fig:Parameter Analysis} (d) shows the detection accuracy performance according to changes of the causal threshold (0.02 to 0.10). Our model achieved the best F1 scores on most datasets at a causal threshold of around 0.06-0.08. 

\begin{table*}[t]
\caption{Comparison of training and inference times}
\label{tab:overhead}
\centering
\begin{tabular}{crrrrrrrr} 
\hline
\multirow{2}{*}{Method} & \multicolumn{2}{c}{SWaT}                                                                                & \multicolumn{2}{c}{HAI}                                                                                & \multicolumn{2}{c}{SMD}                                                                                & \multicolumn{2}{c}{PSM}                                                                                \\ 
\cline{2-9}
                        & \multicolumn{1}{c}{Training}                & \multicolumn{1}{c}{Inference}                    & \multicolumn{1}{c}{Training}                & \multicolumn{1}{c}{Inference}            & \multicolumn{1}{c}{Training}                & \multicolumn{1}{c}{Inference}          & \multicolumn{1}{c}{Training}               & \multicolumn{1}{c}{Inference}           \\ 
\hline
DAGMM                   & {\cellcolor[rgb]{0.447,0.588,0.992}}40.56s  & {\cellcolor[rgb]{0.447,0.588,0.992}}11.86s       & {\cellcolor[rgb]{0.447,0.588,0.992}}45.04s  & {\cellcolor[rgb]{0.447,0.588,0.992}}11.83s     & {\cellcolor[rgb]{0.447,0.588,0.992}}57.59s  & {\cellcolor[rgb]{0.447,0.588,0.992}}15.96s      & {\cellcolor[rgb]{0.447,0.588,0.992}}98.14s  & {\cellcolor[rgb]{0.992,0.408,0.392}}29.81s     \\
USAD                    & {\cellcolor[rgb]{0.992,0.408,0.392}}47.32s  & {\cellcolor[rgb]{0.992,0.408,0.392}}14.45s  & {\cellcolor[rgb]{0.992,0.408,0.392}}54.97s  & {\cellcolor[rgb]{0.992,0.408,0.392}}14.57s & {\cellcolor[rgb]{0.992,0.408,0.392}}68.07s  & {\cellcolor[rgb]{0.992,0.408,0.392}}19.98s   & {\cellcolor[rgb]{0.992,0.408,0.392}}112.27s & {\cellcolor[rgb]{0.992,0.408,0.392}}35.94s  \\
MAD-GAN                 & {\cellcolor[rgb]{0.992,0.408,0.392}}58.45s  & {\cellcolor[rgb]{0.447,0.588,0.992}}9.67s       & {\cellcolor[rgb]{0.992,0.408,0.392}}65.72s  & {\cellcolor[rgb]{0.447,0.588,0.992}}11.40s     & {\cellcolor[rgb]{0.992,0.408,0.392}}84.17s  & {\cellcolor[rgb]{0.447,0.588,0.992}}13.54s      & {\cellcolor[rgb]{0.992,0.408,0.392}}131.96s & {\cellcolor[rgb]{0.447,0.588,0.992}}24.32s     \\
TranAD                  & {\cellcolor[rgb]{0.404,0.992,0.604}}6.64s   & {\cellcolor[rgb]{0.549,1,0.702}}4.98s        & {\cellcolor[rgb]{0.549,1,0.702}}8.02s       & {\cellcolor[rgb]{0.549,1,0.702}}4.49s        & {\cellcolor[rgb]{0.549,1,0.702}}9.73s       & {\cellcolor[rgb]{0.549,1,0.702}}6.20s     & {\cellcolor[rgb]{0.549,1,0.702}}10.38s      & {\cellcolor[rgb]{0.447,0.588,0.992}}10.23s    \\
GDN                     & {\cellcolor[rgb]{0.404,0.992,0.604}}6.21s   & {\cellcolor[rgb]{0.447,0.588,0.992}}6.59s       & {\cellcolor[rgb]{0.549,1,0.702}}7.33s       & {\cellcolor[rgb]{0.549,1,0.702}}6.51s        & {\cellcolor[rgb]{0.447,0.588,0.992}}17.60s  & {\cellcolor[rgb]{0.447,0.588,0.992}}17.30s     & {\cellcolor[rgb]{0.549,1,0.702}}6.61s       & {\cellcolor[rgb]{0.549,1,0.702}}8.87s         \\
MTAD-GAT                & {\cellcolor[rgb]{0.447,0.588,0.992}}13.81s  & {\cellcolor[rgb]{0.549,1,0.702}}5.89s         & {\cellcolor[rgb]{0.447,0.588,0.992}}17.14s  & {\cellcolor[rgb]{0.447,0.588,0.992}}6.81s      & {\cellcolor[rgb]{0.447,0.588,0.992}}17.08s  & {\cellcolor[rgb]{0.549,1,0.702}}7.47s        & {\cellcolor[rgb]{0.447,0.588,0.992}}24.81s  & {\cellcolor[rgb]{0.549,1,0.702}}6.39s      \\
FuSAGNet                & {\cellcolor[rgb]{0.447,0.588,0.992}}9.65s   & {\cellcolor[rgb]{0.992,0.408,0.392}}15.36s & {\cellcolor[rgb]{0.447,0.588,0.992}}11.97s  & {\cellcolor[rgb]{0.992,0.408,0.392}}16.28s   & {\cellcolor[rgb]{0.549,1,0.702}}11.76s      & {\cellcolor[rgb]{0.992,0.408,0.392}}22.33s & {\cellcolor[rgb]{0.549,1,0.702}}17.11s      & {\cellcolor[rgb]{0.447,0.588,0.992}}26.54s   \\
DuoGAT                  & {\cellcolor[rgb]{0.992,0.408,0.392}}46.25s  & {\cellcolor[rgb]{0.992,0.408,0.392}}24.42s      & {\cellcolor[rgb]{0.992,0.408,0.392}}79.53s  & {\cellcolor[rgb]{0.992,0.408,0.392}}26.21s      & {\cellcolor[rgb]{0.992,0.408,0.392}}61.69s  & {\cellcolor[rgb]{0.992,0.408,0.392}}37.97s    & {\cellcolor[rgb]{0.992,0.408,0.392}}139.21s & {\cellcolor[rgb]{0.992,0.408,0.392}}32.02s     \\ 
\hline
CDRL4AD                 & {\cellcolor[rgb]{0.549,1,0.702}}9.43s       & {\cellcolor[rgb]{0.549,1,0.702}}4.41s          & {\cellcolor[rgb]{0.549,1,0.702}}11.49s      & {\cellcolor[rgb]{0.549,1,0.702}}4.83s       & {\cellcolor[rgb]{0.549,1,0.702}}13.43s      & {\cellcolor[rgb]{0.549,1,0.702}}7.83s        & {\cellcolor[rgb]{0.447,0.588,0.992}}22.33s  & {\cellcolor[rgb]{0.549,1,0.702}}7.59s         \\ 
\hline
\multirow{2}{*}{Method} & \multicolumn{2}{c}{SMAP}                                                                                & \multicolumn{2}{c}{MSL}                                                                                & \multicolumn{2}{c}{WADI}                                                                               & \multicolumn{2}{c}{GCP}                                                                                \\ 
\cline{2-9}
                        & \multicolumn{1}{c}{Training}                & \multicolumn{1}{c}{Inference}           & \multicolumn{1}{c}{Training}                & \multicolumn{1}{c}{Inference}             & \multicolumn{1}{c}{Training}                & \multicolumn{1}{c}{Inference}              & \multicolumn{1}{c}{Training}                & \multicolumn{1}{c}{Inference}        \\ 
\hline
DAGMM                   & {\cellcolor[rgb]{0.549,1,0.702}}12.48s      & {\cellcolor[rgb]{0.549,1,0.702}}5.19s         & {\cellcolor[rgb]{0.549,1,0.702}}4.26s       & {\cellcolor[rgb]{0.549,1,0.702}}0.98s      & {\cellcolor[rgb]{0.549,1,0.702}}3.82s       & {\cellcolor[rgb]{0.549,1,0.702}}0.70s        & {\cellcolor[rgb]{0.549,1,0.702}}8.53s       & {\cellcolor[rgb]{0.549,1,0.702}}2.94s         \\
USAD                    & {\cellcolor[rgb]{0.447,0.588,0.992}}13.88s  & {\cellcolor[rgb]{0.447,0.588,0.992}}8.01s    & {\cellcolor[rgb]{0.447,0.588,0.992}}15.17s  & {\cellcolor[rgb]{0.549,1,0.702}}2.64s      & {\cellcolor[rgb]{0.992,0.408,0.392}}68.02s  & {\cellcolor[rgb]{0.549,1,0.702}}1.44s     & {\cellcolor[rgb]{0.447,0.588,0.992}}13.53s  & {\cellcolor[rgb]{0.549,1,0.702}}3.36s      \\
MAD-GAN                 & {\cellcolor[rgb]{0.992,0.408,0.392}}132.21s & {\cellcolor[rgb]{0.447,0.588,0.992}}18.09s      & {\cellcolor[rgb]{0.992,0.408,0.392}}150.17s & {\cellcolor[rgb]{0.992,0.408,0.392}}20.32s    & {\cellcolor[rgb]{0.992,0.408,0.392}}184.21s & {\cellcolor[rgb]{0.992,0.408,0.392}}36.87s      & {\cellcolor[rgb]{0.992,0.408,0.392}}198.72s & {\cellcolor[rgb]{0.992,0.408,0.392}}50.61s      \\
TranAD                  & {\cellcolor[rgb]{0.549,1,0.702}}10.28s      & {\cellcolor[rgb]{0.549,1,0.702}}5.47s          & {\cellcolor[rgb]{0.447,0.588,0.992}}8.22s   & {\cellcolor[rgb]{0.447,0.588,0.992}}7.71s    & {\cellcolor[rgb]{0.447,0.588,0.992}}22.79s  & {\cellcolor[rgb]{0.447,0.588,0.992}}5.70s    & {\cellcolor[rgb]{0.447,0.588,0.992}}29.28s  & {\cellcolor[rgb]{0.549,1,0.702}}7.15s         \\
GDN                     & {\cellcolor[rgb]{0.447,0.588,0.992}}16.23s  & {\cellcolor[rgb]{0.992,0.408,0.392}}47.01s        & {\cellcolor[rgb]{0.549,1,0.702}}8.10s       & {\cellcolor[rgb]{0.992,0.408,0.392}}10.69s       & {\cellcolor[rgb]{0.549,1,0.702}}12.15s      & {\cellcolor[rgb]{0.447,0.588,0.992}}4.23s         & {\cellcolor[rgb]{0.549,1,0.702}}9.35s       & {\cellcolor[rgb]{0.447,0.588,0.992}}11.15s        \\
MTAD-GAT                & {\cellcolor[rgb]{0.992,0.408,0.392}}25.23s  & {\cellcolor[rgb]{0.447,0.588,0.992}}31.62s       & {\cellcolor[rgb]{0.992,0.408,0.392}}17.44s  & {\cellcolor[rgb]{0.447,0.588,0.992}}10.59s     & {\cellcolor[rgb]{0.447,0.588,0.992}}42.79s  & {\cellcolor[rgb]{0.992,0.408,0.392}}7.02s     & {\cellcolor[rgb]{0.992,0.408,0.392}}32.40s  & {\cellcolor[rgb]{0.447,0.588,0.992}}13.71s    \\
FuSAGNet                & {\cellcolor[rgb]{0.549,1,0.702}}5.43s       & {\cellcolor[rgb]{0.549,1,0.702}}1.36s        & {\cellcolor[rgb]{0.549,1,0.702}}3.12s       & {\cellcolor[rgb]{0.549,1,0.702}}0.82s      & {\cellcolor[rgb]{0.549,1,0.702}}5.13s       & {\cellcolor[rgb]{0.549,1,0.702}}3.44s      & {\cellcolor[rgb]{0.549,1,0.702}}6.13s       & {\cellcolor[rgb]{0.992,0.408,0.392}}25.35s  \\
DuoGAT                  & {\cellcolor[rgb]{0.992,0.408,0.392}}92.93s  & {\cellcolor[rgb]{0.992,0.408,0.392}}171.62s     & {\cellcolor[rgb]{0.992,0.408,0.392}}55.99s  & {\cellcolor[rgb]{0.992,0.408,0.392}}41.05s      & {\cellcolor[rgb]{0.992,0.408,0.392}}120.73s & {\cellcolor[rgb]{0.992,0.408,0.392}}15.50s    & {\cellcolor[rgb]{0.992,0.408,0.392}}117.71s & {\cellcolor[rgb]{0.992,0.408,0.392}}72.71s \\ 
\hline
CDRL4AD                 & {\cellcolor[rgb]{0.447,0.588,0.992}}21.83s  & {\cellcolor[rgb]{0.992,0.408,0.392}}37.50s    & {\cellcolor[rgb]{0.447,0.588,0.992}}15.84s  & {\cellcolor[rgb]{0.447,0.588,0.992}}9.77s   & {\cellcolor[rgb]{0.447,0.588,0.992}}56.34s  & {\cellcolor[rgb]{0.447,0.588,0.992}}4.84s   & {\cellcolor[rgb]{0.447,0.588,0.992}}24.09s  & {\cellcolor[rgb]{0.447,0.588,0.992}}17.10s   \\
\hline
\end{tabular}
\begin{flushleft}
\footnotesize{* The unit for both training time per epoch and inference time is in seconds~(s)).}\\
\end{flushleft}
\end{table*}

\subsection{Time Complexity}
To report the scalability of CDRL4AD and state-of-the-art models, we evaluated training time and inference time as key metrics. The training time was determined as the average time per epoch, while the inference time was measured as the total elapsed time required to predict all test samples. 

Table \ref{tab:overhead} shows the performance comparison of training and inference times across all datasets, compared to the state-of-the-art methods. To enable fair time comparisons across different contexts by avoiding the pitfalls of rigid thresholds, we used tertiles. Especially, we categorized time groups as short (in green), moderate (in blue), and long (in red). 

The results indicate that CDRL4AD achieved relatively short inference times across most datasets, with some cases exhibiting the fastest performance. This suggests its high applicability in real-time anomaly detection scenarios. In terms of training time, CDRL4AD generally maintained a moderate level, although it showed slightly higher values for certain datasets (e.g., SMAP and GCP). However, it still demonstrated a balanced performance compared to other models. In contrast, models such as MAD-GAN and DuoGAT exhibited significantly longer training and inference times, which makes them less suitable for real-time applications. Meanwhile, GDN and TranAD attained relatively short training and inference times, which provides efficient computational performance. The reason is that CDRL4AD not only maintains high anomaly detection accuracy but also offers efficient computational performance. As a result, it is well-suited for real-time anomaly detection and batch learning environments.

\begin{figure}[t]
\centering
\includegraphics[width=0.6\columnwidth]{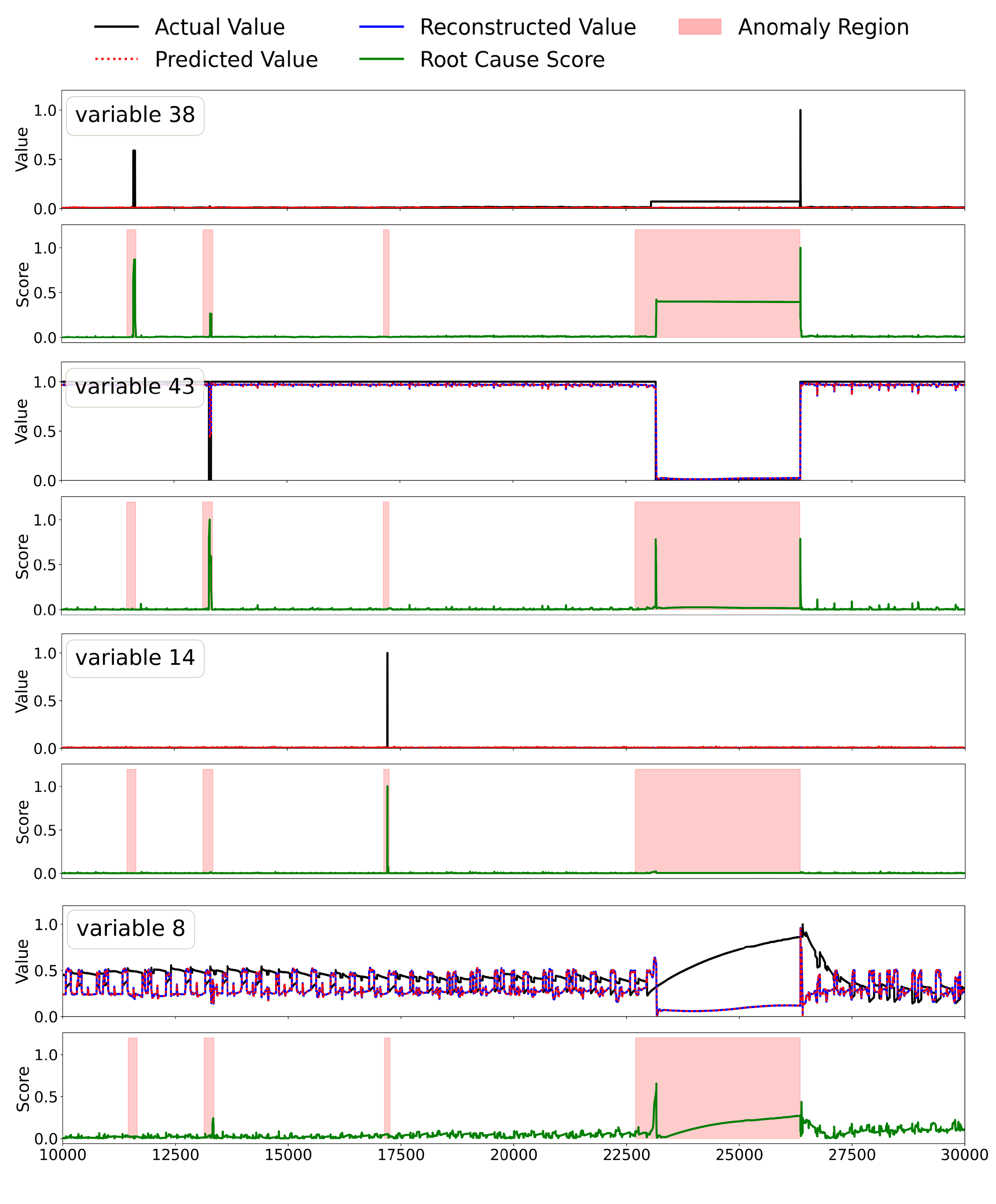}
\caption{Example of anomaly diagnosis process}
\label{fig:anomaly diagnosis}
\end{figure} 

\begin{figure}[t]
\centering
\includegraphics[width=0.6\columnwidth]{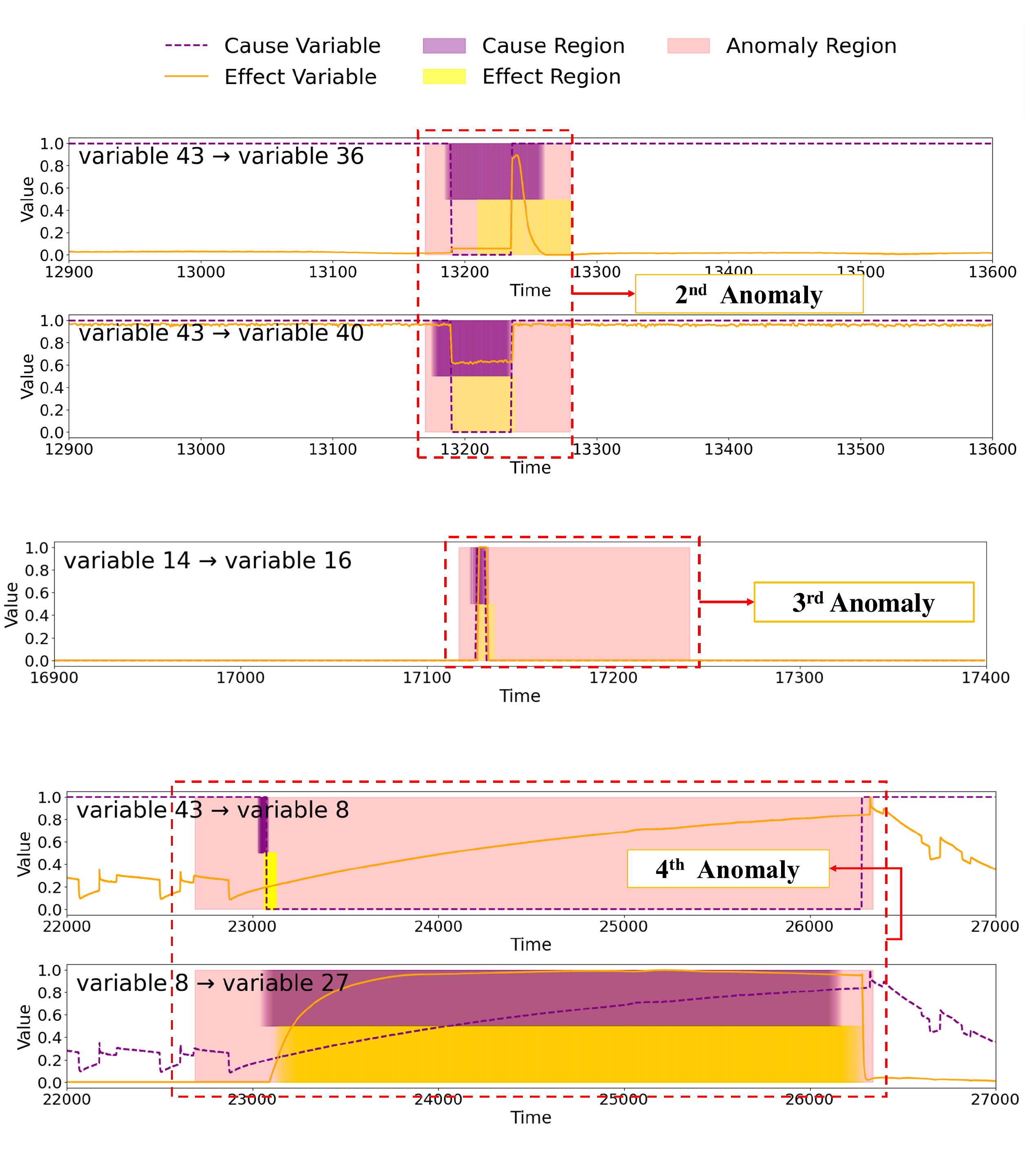}
\caption{Example of causal discovery process}
\label{fig:causal discovery}
\end{figure} 

\section{Case study} \label{case_study}
The deployment of detection models requires a high degree of interpretability, which reflects how well the detection results can be understood and examined by humans \cite{cai2021arm}. In this section, we showcase how our model collaborates with domain experts to identify the root cause of anomalies and discover causal relationships on real-world dataset. To achieve this, we present the anomaly diagnosis and causal discovery processes. 

\subsubsection{\textbf{Anomaly Diagnosis Process}}
This process assists domain experts to diagnose anomalous behaviors in MTS data. As the real-world datasets, a representative portion of SWaT dataset, widely used for diagnosing anomalies \cite{DuoGAT, GReLeN, FuSAGNet}, was extracted. It spanned the time interval from 10,000 to 30,000, which contains the four ground truth anomaly regions. The domains expert assesses the actual values of a detected anomaly event to determine whether a true anomaly has occurred and identifies the variables responsible for it. To achieve this, the expert examines all variables by considering both their actual values and changes to make an informed judgment on anomalies. Consequently, the workload of domain experts can be alleviated by systematically identifying and presenting candidate variables that may have caused the anomaly.

Figure \ref{fig:anomaly diagnosis} shows four examples of anomaly diagnosis process examined by domain experts. He/she selected four variables as potential primary root causes in the same four anomaly regions. Each variable is represented by two subgraphs. The upper subgraph shows the discrepancy between the actual (in \textit{black} line), predicted (in \textit{red dotted} line), and reconstructed values (in \textit{blue} line) of each variable. The lower subgraph quantifies the anomaly severity through shows the root cause scores (in \textit{green} line).

Specifically, variables 38, 43, 14, and 8 were identified as the 1st ranked root cause candidates for the first to fourth anomalies, respectively. For example, the variable 8 was identified as the 2nd ranked candidate for the fourth anomaly. Besides, the experts easily found that all first candidate variables exhibited strong deviations from their normal trends, along with increased root cause scores. The experts determined variables with higher root cause scores as the primary causes in each anomaly region. Hereafter, the notation \(X_{i}\) in Table~\ref{tab:1} is used to indicate each variable. As a result, variables \(X_{38}\), \(X_{43}\), \(X_{14}\), and \(X_{43}\) were identified as the root causes of the corresponding anomaly regions by the domain experts.

\subsubsection{\textbf{Causal Discovery Process}}
This process supports humans predict time-lagged causal relationships through causal discovery. Figure \ref{fig:causal discovery} shows an example of five time-lagged cause-and-effect relationships from the root causes of the second, third, and fourth anomaly regions in Figure \ref{fig:anomaly diagnosis}. In each subfigure, the cause and effect variables are represented by \textit{purple dashed} and \textit{orange solid} lines, respectively. The cause and effect regions are depicted by \textit{purple} and \textit{yellow boxes}, respectively. 

Note that the cause and effect variable pairs were automatically discovered by our model. Besides, the cause and effect regions were determined based on the continuity of the relationship between the discovered pair.

First of all, the third subfigure shows a simple cause-and-effect relationship. The domain expert observed that the value of \(X_{16}\) deviated from its normal trend immediately after the value of \(X_{14}\) showed abnormal changes. It is worth noting that the cause region precedes the effect region by multiple time steps. In other words, our model found time-lagged causal relationships between \(X_{14}\) and \(X_{16}\). The reason was that the cause and effect regions are smaller than the anomaly regions, which results in some space between them. This gap exists because the ground truth anomaly labels were manually assigned by humans after the anomalies had occurred.
 
Notably, the domain expert identified complex causal structures involving \(X_{8}\), \(X_{27}\), \(X_{36}\), \(X_{40}\), and \(X_{43}\). As shown in the first and second subfigures, he/she observed that the root cause \(X_{43}\) of the second anomaly serves as a common cause variable for both \(X_{36}\) and \(X_{40}\)~(\( X_{36} \leftarrow X_{40} \rightarrow X_{43} \)). Similarly, in the fourth and fifth subfigures, the expert discovered a causal chain formulation~(\( X_{43} \rightarrow X_{8} \rightarrow X_{27} \)). Especially, the \(X_{43}\) triggered the anomalous behavior of \(X_{8}\). Interestingly, the anomalous state in \(X_{8}\) causes the anomaly in \(X_{27}\), with the cause region occurring several time intervals before the effect region. It is worth noting that understanding the causal structures benefits both prediction and interpretation. CDRL4AD plays a crucial role in this process by assisting domain experts identify causal relationships and analyze system behavior.

\section{Conclusion} \label{conclusion}
To conclude, we have proposed CDRL4AD that accurately detects anomalies with CDR for analyze data with latent variables related to causation and correlations in multivariate time series. Especially, we first have constructed a temporal heterogeneous graph that captures time-lagged causal relationships, node and edge correlations, and temporal dependencies within data variations. Second, our approach has identified cause-and-effect variables across different time periods and also disentangled latent variables exhibiting complex causality using VAE with a multi-head decoder. Last, our model has captured feature correlations at both node and edge levels, as well as temporal dependencies. 

In the evaluation, we have demonstrated that CDRL4AD outperformed state-of-the-art methods in terms of detection accuracy and root cause analysis on real-world datasets. Besides, both sensitivity and time complexity analysis have revealed stable and efficient performance of our model according to varying situations. Furthermore, our case study on both anomaly diagnosis and causal discovery processes has assisted domain experts in identifying primary causes of anomalies and distinguishing time-lagged causal relationships.

\bibliographystyle{unsrt}  

\end{document}